\documentclass{article}
\usepackage{amsmath}
\usepackage{arxiv}

\usepackage[utf8]{inputenc} 
\usepackage[T1]{fontenc}    
\usepackage[hidelinks]{hyperref}       
\usepackage{url}            
\usepackage{booktabs}       
\usepackage{amsfonts}       
\usepackage{nicefrac}       
\usepackage{microtype}      
\usepackage{cleveref}       
\usepackage{lipsum}         
\usepackage{graphicx}
\usepackage[square, numbers]{natbib}
\usepackage{mathtools}
\usepackage{enumitem}
\usepackage{leftindex}
\usepackage{doi}
\usepackage{stackengine}
\usepackage{float}
\title{On Fuzzy Cardinal Semantic Transformations}

\date{}

\author{ \href{https://orcid.org/0000-0001-8935-0338}{\hspace{1mm}Alexander Yu. Chunikhin}\\
	Palladin Institute of Biochemistry\\
	NAS of Ukraine\\
	\texttt{alexchunikhin@gmail.com} \\
	ORCID 0000-0001-8935-0338\\
	\And
\href{https://orcid.org/my-orcid?orcid=0000-0003-4034-3393}{\hspace{1mm}Vadym Zhytniuk}\\
Palladin Institute of Biochemistry\\
NAS of Ukraine\\
\texttt{v.zhytniuk@kau.edu.ua} \\
ORCID 0000-0003-4034-3393\\
}

\hypersetup{
pdftitle={A template for the arxiv style},
pdfsubject={q-bio.NC, q-bio.QM},
pdfauthor={David S.~Hippocampus, Elias D.~Striatum},
pdfkeywords={First keyword, Second keyword, More},
colorlinks=false,
}
\newtheorem{theorem}{Definition}
\begin{document}
\maketitle

\begin{abstract}
The concept of fuzzy cardinal semantic transformation as a basis for creating fuzzy semantic numeration systems is introduced in this work. Both fuzziness of the initial data - cardinals of abstract entities - and fuzziness of the parameters of the cardinal semantic operators are considered. We also expressed cardinal semantic transformations for discrete fuzzy numbers and for continuous triangular fuzzy numbers. The principle of formation of the fuzzy common carry in the cardinal semantic operators with multiple inputs is formed.
\end{abstract}

\keywords{cardinal semantic operator \and fuzzy numbers \and fuzzy cardinal semantic transforamtion \and fuzzy common carry}

\section{Introduction}
Modern numeration systems do not quite meet requirements of the problems to be solved so they do not consider variety and multidimensional nature of the numerical input data, multistage and multitask nature of the processes of numerical processing. For instance,  in practice, one of the most crucial and common aspects of numerical data is  its  uncertainty. At present, mathematical theories that are considered to treat uncertainty are interval analysis \cite{methods}, theory of probability and theory of fuzzy sets \cite{Piegat}. \\
A numeration system is a symbolic method of representing numbers using signs. The semantics of the traditional place-value representation can be expressed as follows: $n$ units of some abstract entity $i$ are given the meaning ($\sim$>) of a unit of another abstract entity j: $n1_{i} = n_{i} \sim> 1_{j}$. Assume that there are such abstract entities $i$ whose $n$ units $n_{i}$ are given the meaning of both the unit of the abstract entity $j (1_{i})$ and the unit of the abstract entity $k (1_{k})$ simultaneously: $n_{i} \sim> (1_{j}, 1_{k})$. Consider another situation: to form a unit of an abstract entity $k$, exactly n units of an abstract entity $i$, and m units of an abstract entity $j$, are required: $(n_{i}, m_{j} ) \sim> 1_{k}$. \\
The semantic numeration systems $(SNS)$ with crisp cardinal semantic transformations were developed in the fundamentals \cite{fundamentals}, \cite{SNS}. In current work we consider $SNS$ development  in case of fuzzy  type of uncertainty. Both fuzziness of the initial data - cardinals of abstract entities and possible fuzziness of cardinal semantic operators parameters are considered. All values of discrete and continuous fuzzy numbers are considered to be natural numbers. 

\section{Preliminaries}
Let us remind the main definitions that are related to the cardinal semantic operators \citep{fundamentals}, \citep{SNS}.
\begin{theorem}
	A Cardinal Semantic Operator is a multivalued mapping of the cardinal semantic
	multeity on itself, which associates a set of entity operands from the multeity with a set of entity
	images from the same multeity, transforming their cardinals using the operations defined by the
	operator signature: Signt(CSO) = (K, Form, | n > w, | r > v ), where K is a kind of the operator, Form is
	a type of the operator, | n > is a radix vector, and |r> is a conversion vector. The pair (W, V) is a valence
	of the Cardinal Semantic Operator.
\end{theorem}

Let us define the main forms of cardinal semantic operators of the ($\uparrow$ \#) kind (Carry).
\begin{theorem}
	L-operator (Line-operator): ($\uparrow\#, L, n_{i} , r_{ij} $) $-$ a cardinal semantic operator of valency
	(W, V) = (1, 1), which assigns (gives the meaning of) $r_{ij}$ units of the transformant $q_{j}$ , added to the
	cardinal $N_{j}$ of the abstract entity $\text{C\AE}_{j}$  , to each $n_{i}$ of the cardinal abstract entity $\text{C\AE}_{i}$ .
\end{theorem}
\begin{figure}[h]
	\centering
	\includegraphics[scale=0.3]{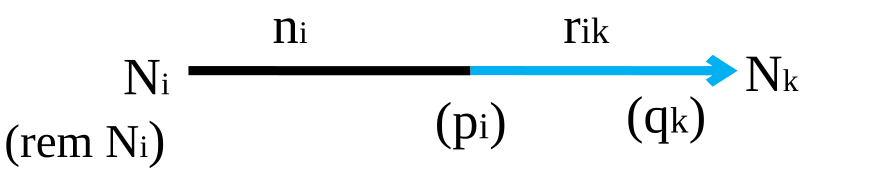}
	\caption{L - operator.}
	\label{fig:fig1}
\end{figure}
When L-operator acts on $\text{C\AE}_{j}$-operand, the following operations are performed:
\begin{enumerate}[label = (\roman*)]
	\item $p_{i} = \big \lfloor \nicefrac{N_{i}}{n_{i}} \big \rfloor  - $ calculation of a carry;
	\item $N_{i}^{'} = rem(N_{i}) = N_{i} - p_{i}n_{i} = N_{i}mod(n_{i}) - $finding the remainder in $\text{C\AE}_{i}$
	\item $q_{j} = p_{i}r_{ij} - $calculation of j-transformant value;
	\item $N_{j}^{'} = N_{j}+q_{j} - $finding the change of the $\text{C\AE}_{j} - $image cardinal. 
\end{enumerate}

\begin{theorem}
	D-operator (Distribution operator): ($\uparrow\#, D, n_{i} , (r_{ij} , . . . , r_{ik} )$) $-$ a cardinal semantic
	operator of valency (W, V) = (1, v), which assigns the following units to each $n_{i}$ of the cardinal
	abstract entity $\text{C\AE}_{i}$ v transformants: $r_{ij}$ units of j-transformants $q_{j}$ for the cardinal abstract entity
	$\text{C\AE}_{j}$ , . . . , and $r_{ik}$ units of k-transformants $q_{k}$ for the cardinal abstract entity $\text{C\AE}_{k}$
\end{theorem}
\begin{figure}[h]
	\centering
	\includegraphics[scale=0.3]{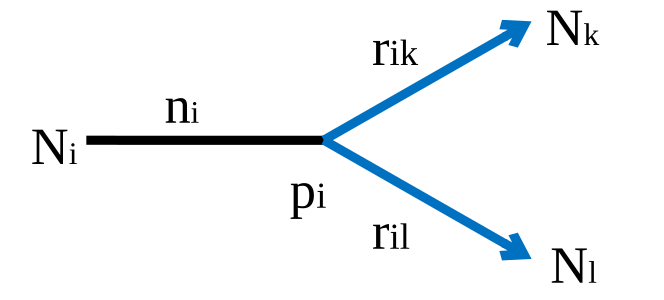}
	\caption{$D_{2}$ - operator.}
	\label{fig:fig1}
\end{figure}
When $D_{2}$-operator acts on $\text{C\AE}_{i} - $operand, the following operations are performed:
\begin{enumerate}[label = (\roman*)]
	\item $p_{i} = \big \lfloor \nicefrac{N_{i}}{n_{i}} \big \rfloor - $calculation of a carry;
	\item $N_{i}^{'} = rem(N_{i}) = N_{i} - p_{i}n_{i} = N_{i}mod(n_{i}) - $finding the remainder in $\text{C\AE}_{i}$;
	\item $q_{l} = p_{i}r_{il}, q_{k} = p_{i}r_{ik} - $calculation of the partial transformants;
	\item $N_{l}^{'} = N_{l} + q_{l}, N_{k}^{'} = N_{k} + q_{k} - $finding the change of the $\text{C\AE} - $images ($\text{C\AE}_{l}$, $\text{C\AE}_{k}$) cardinals.
\end{enumerate}

\begin{theorem}
	F-operator (Fusion operator): ($\uparrow\#, F, (n_{i} , . . . , n_{j} ), r_{.k} $) $-$ a cardinal semantic operator
	of valency (W, V) = (w, 1), which assigns $r_{.k}$ units of the transformant $q_{k}$ to each w-tuple ($n_{i}$ , . . . ,
	$n_{j}$ ) of $\text{C\AE}$-operands for the cardinal abstract entity $\text{C\AE}_{k}$ .
\end{theorem}

\break
\begin{figure}[h]
	\centering
	\includegraphics[scale=0.3]{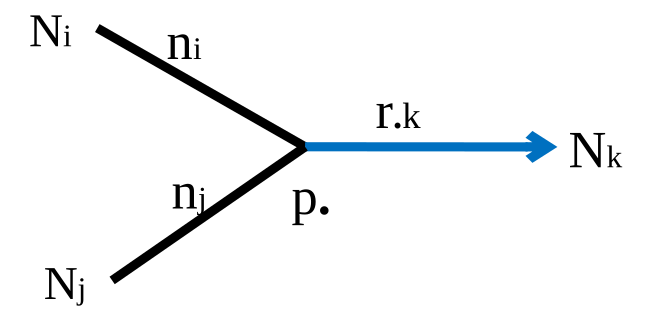}
	\caption{$\leftindex_{2}{F}$ - operator.}
	\label{fig:fig1}
\end{figure}

When $\leftindex_{2}{F} - $operator acts on $\text{C\AE}_{i, j} - $operands, the following operations are performed:
\begin{enumerate}[label = (\roman*)]
	\item $p_{i} = \big \lfloor \nicefrac{N_{i}}{n_{i}} \big \rfloor, p_{j} = \big \lfloor \nicefrac{N_{j}}{n_{j}} \big \rfloor - $calculation of partial carries;
	\item $p_{.} = min(p_{i}, p_{j}) - $calculation of the common carry;
	\item $N_{i}^{'} = N_{i} - p_{.}n_{i}, N_{j}^{'} = N_{j} - p_{.}n_{j} - $calculation of the remainder in $\text{C\AE}_{i}, \text{C\AE}_{j}$;
	\item $q_{k} = p_{.}r_{.k} - $calculation of the transformant;
	\item $N_{k}^{'} = N_{k} + q_{k} - $finding the change of the $\text{C\AE}_{k} - $image cardinal.
\end{enumerate}

\begin{theorem}
	M-operator (Multi-operator): ($\uparrow\#, M, (n_{i} , . . . , n_{j} ), (r_{.k} , . . . , r_{.l} )$) $-$ a cardinal
	semantic operator of valency (W, V) = (w, v), which assigns v-tuple conversion coefficients ($r_{.k} , . . . ,
	r_{.l} $) of transformants to the w-tuple ($n_{i} , . . . , n_{j} $) of $\text{C\AE}$-operands: $r_{.k}$ units of k-transformant $q_{k}$ for the cardinal abstract entity $\text{C\AE}_{k}$ , . . . , and $r_{.l}$ units of l-transformants $q_{l}$ for the cardinal abstract entity $\text{C\AE}_{l}$ .
\end{theorem}
\begin{figure}[h]
	\centering
	\includegraphics[scale=0.3]{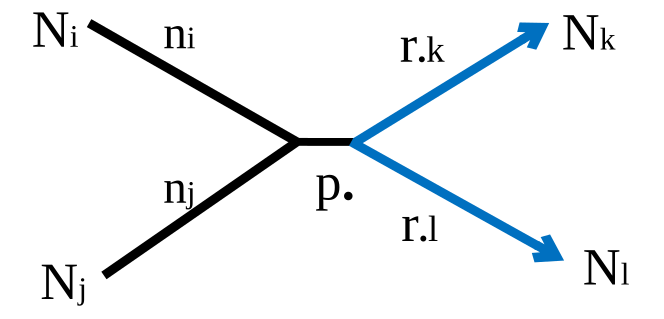}
	\caption{ $\leftindex_{2}{M}_{2}$- operator.}
	\label{fig:fig1} 
\end{figure}

When $\leftindex_{2}{M}_{2} - $operator acts on $\text{C\AE}_{i, j} - $operands, the following operations are performed:
\begin{enumerate}[label = (\roman*)]
	\item $p_{i} = \big \lfloor \nicefrac{N_{i}}{n_{i}} \big \rfloor, p_{l} = \big \lfloor \nicefrac{N_{j}}{n_{j}} \big \rfloor -  $calculation of a carry;
	\item $p_{.} = min(p_{i}, p_{j}) - $calculation of the common carry;
	\item $N_{i}^{'} = N_{i} - p_{.}n_{i}, N_{j}^{'} = N_{j} - p_{.}n_{j} - $calculation of the remainder in $\text{C\AE}_{i}, \text{C\AE}_{j}$;
	\item $q_{k} = p_{.}r_{.k}, q_{l} = p_{.}r_{.l} - $calculation of the partial transformants;
	\item $N_{k}^{'} = N_{k} + q_{k}, N_{l}^{'} = N_{l} + q_{l} - $finding the changes of the $\text{C\AE}_{k, l} - $image cardinals.
\end{enumerate}
These four cardinal semantic operators form the operator basis of any semantic numeration system.
\section{Discrete fuzzy operator}
All algebraic operations on discrete fuzzy numbers are performed in accordance with Zadeh extension principle \cite{Piegat}, so that \\
\begin{equation}
	\begin{cases}
		\forall \tilde{A}  = \sum\limits_{x}^{} [x|\mu_{\tilde{A}}(x)] , \tilde{B} = \sum\limits_{y}[y|\mu_{\tilde{B}}(y)]\\
		\exists \tilde{C} = \tilde{A} \circ \tilde{B} = \sum\limits_{z}[z|\mu_{\tilde{C}}(z)]\\ 
		z = x \circ y\\
		\mu_{\tilde{C}} (z)= \bigvee\limits_{x \circ y}[\mu_{\tilde{A}}(x)\wedge \mu_{\tilde{B}}(y)]
	\end{cases}
\end{equation}

In principle, the following variances of the initial fuzziness are possible:
\begin{enumerate}[label = (\roman*)]
	\item "Fuzzy input, crisp operator". In this case, at least, one C\AE-operand has a fuzzy cardinal: 
	\[card(\text{C\AE}_{i}) = \tilde{N}_{i} \]
	\item "Crisp input, fuzzy operator". All the $card({N}_{i})$ are crisp. Some of the parameters of $CSO (n, r)$, or all of them, are fuzzy. 
	\item "Fuzzy input, fuzzy operator". 
\end{enumerate}
\subsection{Fuzzy input, crisp L-operator}
\[\tilde{N}_{i} \xmapsto[p_{i}]{n_{i}  \hspace{1cm} r_{ij}} q_{j} \xmapsto[]{}N_{j}\]
In this paragraph, we consider the initial cardinal $\tilde{N}_{i}$ to be fuzzy and all components of the cardinal semantic operator are crisp. The initial cardinal $\tilde{N}_{i}$ is defined as
\begin{equation*}
	\hat{N}_{i} = \sum_{t}[t|\mu_{\hat{N}_{i}}(t)], t \in \mathbb{N}, \mu_{\hat{N}_{i}} \in (0,1).
\end{equation*}

If, at least, one operand during execution of the operation is fuzzy, than the result, due to the necessity,  is fuzzy as well. When the crisp cardinal semantic operator acts on fuzzy cardinals, the following operations are performed:
\begin{enumerate}[label = (\roman*)]
	\item $\hat{p}_{i} = \bigg \lfloor  \frac{\hat{N}_{i}}{n_{i}}\bigg \rfloor  = \sum\limits_{\big \lfloor  \nicefrac{t}{n_{i}}\big \rfloor}[(\bigg \lfloor \frac{t}{n_{i}} \bigg \rfloor )|\mu_{\hat{N}_{i}}(g)]  = \sum\limits_{g}[g|\mu_{\hat{p}_{i}}(g)] - $calculation of a carry,\\
	where $g$ is defined as \[g = \bigg \lfloor \frac{t}{n_{i}} \bigg \rfloor, g \in  \mathbb{N}\]
	\item $\hat{q}_{j} = \hat{p}_{i}r_{ij} = r_{ij} \sum\limits_{g}[g|\mu_{\hat{p}_{i}}(g)] = \sum\limits_{gr_{ij}}[gr_{ij}|\mu_{\hat{q}_{i}}(gr_{ij})] = \sum\limits_{\varphi}[\varphi|\mu_{\hat{q}_{j}}(\varphi)] - $calculation of the transformant
	\item  $\hat{N}_{j}^{'} = \overset{?}{N}_{j} + \hat{q}_{j} - $calculation of the image cardinal. \\
	As the initial cardinal ($\overset{?}{N} _{j}$) may be either crisp or fuzzy, there are two ways of its calculation. 
\end{enumerate}

\begin{enumerate}
	\item If the image cardinal is crisp, the following operation is performed:
	\[\hat{N}_{j}^{'} = N_{j} + \sum_{\varphi}[\varphi|\mu_{\hat{q}_{j}})\varphi] = \sum_{\varphi+N_{j}}[\varphi+N_{j}|\mu_{\hat{N}_{j}^{'}}(\varphi+N_{j})]\]
	\item If the initial image cardinal is fuzzy and expressed as
	\[\hat{N}_{j} = \sum_{s}[s|\mu_{\hat{N}_{j}}(s)],\]
	the following operation performs:
	\[\hat{N}_{j}^{'} = \sum_{s}[s|\mu_{\hat{N}_{j}}(s)] + \sum_{\varphi}[\varphi|\mu_{\hat{q}_{j}}(\varphi)] = \sum_{s + \varphi}[s+\varphi|\mu_{\hat{N}_{j}^{'}}(s+\varphi)]\]
\end{enumerate}
		
\begin{enumerate}[label = (\roman*)]
	\setcounter{enumi}{3}
	\item $Rem(\hat{N}_{i}) = \hat{N}_{i} - \hat{p}_{i}n_{i} =\sum\limits_{t}[t|\mu_{\hat{N}_{i}}(t)] - \sum\limits_{g}[g|\mu_{\hat{p}}(g)] = \sum\limits_{t-g}[(t-g)|\mu_{\hat{N}_{i}^{'}}(t-gn_{i})] - $calculation of the remainder.
\end{enumerate}

\subsection{Crisp input, fuzzy L-operator}
If a cardinal semantic operator is fuzzy, there are three possible ways of it's performance, when the radix is fuzzy with crisp conversion rate, and when the radix is crisp with fuzzy conversion rate, and when both the radix and the conversion rate are fuzzy.
\begin{enumerate}
	\item The radix is fuzzy and defined as:
	\[\hat{n}_{i} = \sum_{s}[s|\mu_{\hat{n}_{i}}(s)],\]
	the following operations perform:
	\begin{enumerate}[label = (\roman*)]
		\item $\hat{p}_{i} = \big \lfloor \frac{N_{i}}{\hat{n}_{i}} \big \rfloor  = \bigg \lfloor \frac{N_{i}}{\sum\limits_{s}[s|\mu_{\hat{n}_{i}}(s)]} \bigg \rfloor = \sum\limits_{\nicefrac{N_{i}}{s}}[\frac{N_{i}}{s}|\mu_{\hat{p}_{i}}(\big \lfloor \frac{N_{i}}{s} \big \rfloor )] = \sum\limits_{g}[g|\mu_{\hat{p}_{i}}(g)] - $calculation of a carry;
		\item $\hat{q}_{j} = \hat{p}_{i}r_{ij} = \sum\limits_{\varphi}[\varphi|\mu_{\hat{q}_{j}}(\varphi)] - $calculation of a transformant;
		\item $\hat{N}_{j}^{'} = \hat{N}_{j} + \hat{q}_{j} = \hat{N}_{j} + \sum\limits_{\varphi}[\varphi|\mu_{\hat{q}_{j}}(\varphi)] = \sum\limits_{\hat{N}_{j}^{'} + \varphi}[\hat{N}_{j} + \varphi|\mu_{\hat{N}_{j}^{'}}(\hat{N}_{j}^{'} +\varphi)]$
		\item $\hat{N}_{i}^{'} = Rem(N_{i})  = N_{i} - \hat{p}_{i}|\hat{n}_{i} = N_{i} - \sum\limits_{g}[g|\mu_{\hat{p}_{i}}(g)]\sum\limits_{s}[s|\mu_{\hat{n}_{i}}(s)] = N_{i} - \sum\limits_{gs}[gs|\mu_{\hat{p}_{i}\hat{n}_{i}}(gs)] = \\
		= \sum\limits_{N_{i}-gs}[(N_{i}-gs)|\mu_{\hat{N}_{i}^{'}}(N_{i}-gs)] = \sum\limits_{\psi}[\psi|\mu_{\hat{N}_{i}^{'}}(\psi)] - $calculation of a remainder.
	\end{enumerate}

	\item The conversion rate is fuzzy and defined as
	\[\hat{r}_{ij} = \sum_{\nu}[\nu|\mu_{r_{ij}}(\nu)],\]
	the following operations perform:
	\begin{enumerate}[label = (\roman*)]
		\item $p_{i} = \big \lfloor \frac{N_{i}}{n_{i}} \big \rfloor  - $calculation of a carry;
		\item $\hat{q}_{j} = p_{i}\hat{r}_{ij} = p_{i}\sum\limits_{\nu}[\nu|\mu_{\hat{r}_{ij}}(\nu)] = \sum\limits_{p_{i}\nu}[p_{i}\nu|\mu_{\hat{q}_{j}}(p_{i}\nu)] - $calculation of a conversion rate;
		\item $\hat{N}_{j}^{'} = \hat{N}_{j} + \hat{q}_{j} = \hat{N}_{j} + \sum\limits_{p_{i}\nu}[p_{i}\nu|\mu_{\hat{q}_{j}}(p_{i}\nu)]  = \sum\limits_{\hat{N}_{j} + p_{i}\nu}[\hat{N}_{j} + p_{i}\nu|\mu_{\hat{N}_{j}}(\hat{N}_{j} + p_{i}\nu)] $
		\item $N_{i}^{'} = Rem(N_{i}) = N_{i} - p_{i}n_{i} - $calculation of a remainder
	\end{enumerate}

	\item If both conversion rate and an a radix are fuzzy, the following operation perform:
	\begin{enumerate}[label = (\roman*)]
		\item $\hat{p}_{i} = \big \lfloor \frac{N_{i}}{\hat{n}_{i}} \big \rfloor  = \bigg \lfloor \frac{N_{i}}{\sum\limits_{s}[s|\mu_{\hat{n}_{i}}(s)]} \bigg \rfloor = \sum\limits_{\nicefrac{N_{i}}{s}}[\frac{N_{i}}{s}|\mu_{\hat{p}_{i}}(\big \lfloor \frac{N_{i}}{s} \big \rfloor ) = \sum\limits_{g}[g|\mu_{\hat{p}_{i}}(g)] - $calculation of a carry;
		\item $\hat{q}_{j} = \hat{p}_{i}\hat{q}_{j} = \sum\limits_{g}[g|\mu_{\hat{p}_{i}}(g)]  \sum\limits_{\nu}[\nu|\mu_{r_{ij}}(\nu)]  = \sum\limits_{g\nu}[g\nu|\mu_{\hat{q}_{j}}(g\nu)] - $calculation of a transformant;
		\item $\hat{N}_{j}^{'} = \hat{N}_{j} + \hat{q}_{j} = \hat{N}_{j} +  \sum\limits_{g\nu}[g\nu|\mu_{\hat{q}_{j}}(g\nu)] = \sum\limits_{\hat{N}_{j} + g\nu}[\hat{N}_{j} + g\nu|\mu_{\hat{N}_{j}}(\hat{N}_{j} + g\nu)] $
		\item $\hat{N}_{j}^{'} = \hat{N}_{j} + \hat{q}_{j} = \hat{N}_{j} + \sum\limits_{g\nu}[g\nu|\mu_{\hat{q}_{j}}(g\nu)] = \sum\limits_{\hat{N}_{j} + g\nu}[\hat{N}_{j} + g\nu|\mu_{\hat{N}_{j}}(\hat{N}_{j} + g\nu)] - $calculation of a remainder;
	\end{enumerate}
\end{enumerate}
\subsection{Fuzzy input, fuzzy L-operator}
In this paragraph we consider both the cardinal semantic operator and  the input to be fuzzy. 
\[\hat{N}_{i} \xmapsto[\hat{p}_{i}]{\hat{n}_{i}  \hspace{1cm} \hat{r}_{ij}} \hat{q}_{j} \xmapsto[]{}\hat{N}_{j}\]
When fuzzy semantic operator acts on fuzzy cardinal, the following operations are performed:
\begin{enumerate}[label = (\roman*)]
	\item $\hat{p}_{i} = \big \lfloor \frac{\hat{N}_{i}}{\hat{n}_{i}} \big \rfloor = \bigg \lfloor \frac{\sum\limits_{t}[t|\mu_{\hat{N}_{i}}(t)]}{\sum\limits_{s}[s|\mu_{\hat{n}_{i}}(s)]} \bigg \rfloor = \sum\limits_{\big \lfloor \nicefrac{t}{s}\big \rfloor }[\big \lfloor \frac{t}{s} \big \rfloor |\mu_{\hat{p}_{i}}(\big \lfloor \frac{t}{s} \big \rfloor )] = \sum\limits_{g}[g|\mu_{\hat{p}_{i}}(g)] - $calculation of a carry,\\
    where $g$ is defined as \[g = \bigg \lfloor \frac{t}{s} \bigg \rfloor ;\]
	\item $\hat{q}_{j} = \hat{p}_{i}\hat{r}_{ij} = \sum\limits_{g}[g|\mu_{\hat{p}_{i}}(g)]\sum\limits_{\nu}[\nu|\mu_{\hat{r}_{ij}}(\nu)] = \sum\limits_{g\nu}[g\nu|\mu_{\hat{q}_{j}}(g\nu)] - $calculation of a transformant;
	\item Calculation of an image cardinal is analogous as in the paragraph 3.2;
	\item $\hat{N}_{i}^{'} = Rem(\hat{N}_{i}) = \sum\limits_{t}[t|\mu_{\hat{N}_{i}}(t)] - \sum\limits_{g}[g|\mu_{\hat{p}_{i}}(g)]\sum\limits_{s}[s|\mu_{\hat{n}_{i}}(s)] = \sum\limits_{t-gs}[t-gs|\mu_{\hat{N}_{i}^{'}}(t-gs)] - $calculation of a remainder.
\end{enumerate}

As a result, we described all algebraic manipulations that are performed by cardinal semantic operator with different initial fuzziness: fuzzy inputs, fuzzy operator and the whole fuzziness. 
\section{The principle of common carry formation in CSO of \boldmath$\leftindex_{W}{F}_{V}$ and \boldmath$\leftindex_{W}{M}_{V}$ types}
Unlike the comparison of two fuzzy numbers and	selection of a fuzzy number that is less than the other, in CSO ($\leftindex_{W}{F}_{V}$ and $\leftindex_{W}{M}_{V}$),  common carry has to be 'synthesized' based on a new set of partial carries (${p_{i},...,p_{j}}$). Forming, but not choosing!\\
The concept of common carry calculation lies in its formation from the partial carries of each $\text{C\AE}$-operand of $CSO$, with guarantee that the cardinals of the remainders are zeroes or natural numbers in each $\text{C\AE}$ of an operand.  \\
Let  $\tilde{A}$ be a continuous triangular fuzzy number and its membership ($\mu_{\tilde{A}}$) function is defined as in \cite{Piegat}:
	\begin{equation}
	\mu_{\tilde{A}}(x) = \begin{cases}
		0, & x < \underline{a}; \\
		\frac{x - \underline{a}}{a - \underline{a}}\, &  \underline{a} \leq x \leq a; \\
		\frac{\bar{a} - x}{\bar{a} - a}, &  a \leq x \leq \bar{a};\\
		0, &  x > \bar{a}
	\end{cases} \hspace{2cm} \forall x \; \in \; X.
\end{equation}
Where $x$ is an element of a support ($X$).\\ So each triangular fuzzy number may be defined as a triple ($\underline{a}_{\tilde{A}}; a_{\tilde{A}}; \bar{a}_{\tilde{A}}$), where $\underline{a}_{\tilde{A}}; a_{\tilde{A}}; \bar{a}_{\tilde{A}}$ are a left bound, a mode and a right bound of an arbitrary fuzzy number $\tilde{A}$. In case that partial carries are represented as such continuous triangular fuzzy numbers, the following "min"- method of forming the fuzzy common carry is proposed. 

\begin{equation}
	\tilde{p}_{.} = (\underline{p}_{.}; p_{.}; \bar{p}_{.}) ,
\end{equation}
where $\underline{p}, p, \bar{p}$ are a lower bound, a mode and an upper bound of the fuzzy common carry $p_{.}$. They are defined as the minimum value of bounds and the mode of the partial carries, so that 
\begin{equation}
	\underline{p} = min(\underline{p}_{i},...,\underline{p}_{j})
\end{equation}
\begin{equation}
	mode(p) = min[mode(p_{i}),...,mode(p_{j})]
\end{equation}
\begin{equation}
	\bar{p} = min(\bar{p}_{i},...,\bar{p}_{j})
\end{equation}

Formation of the common carry from partial carries is illustrated in fig.5, where $\tilde{p}_{i}, \tilde{p}_{j}, \tilde{p}_{k}$ are partial carries and $\tilde{p}_{o}$ is the common carry. Linear ordering of the components $\underline{a}; a; \bar{a}$ ensures the uniqueness of the solution $\tilde{p} = form(\tilde{p}_{i},...,\tilde{p}_{j})$.
\begin{figure}[H]
	\centering
	\includegraphics[scale=0.6]{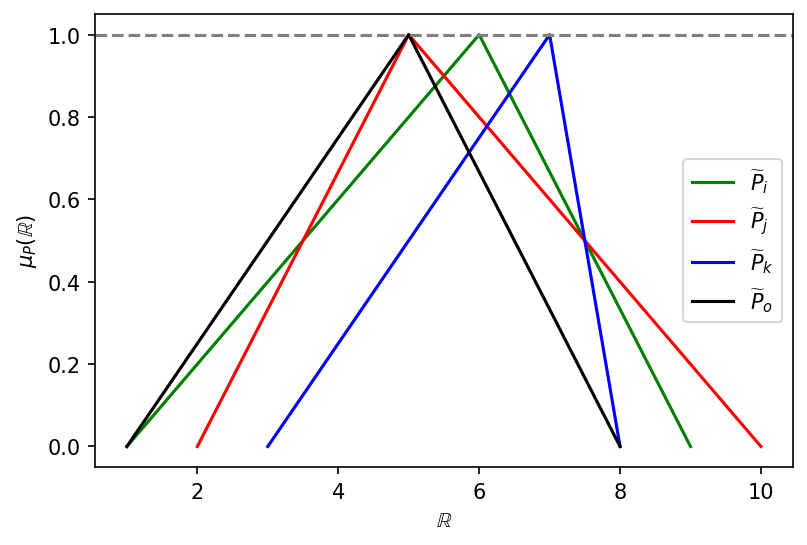}
	\caption{Formation of a fuzzy common carry.}
	\label{fig:fig1}
\end{figure}

The less obvious approach in common carry formation is when partial carries are discrete fuzzy numbers of a type \[\hat{A} = \sum\limits_{t}[t|\mu_{\hat{A}}(t)],\] where $t$ is an element of the support of a fuzzy number $\hat{A}$ and $\mu_{\hat{A}}$ is a corresponding values of the membership function. In the most simple case, when the supports of the partial carries do not intersect, the resulting common carry is a partial carry with the least value of the mode.
\begin{equation}
	\hat{p} = [\hat{p}_{i}(t)|\mu_{\hat{p}_{i}}(t) = 1]
\end{equation}

Let us consider common carry formation in case of $CSO$ with $\leftindex_{2}{F}$ and fuzzy discrete inputs ($\text{C|AE}$-operand). Let an initial cardinal be defined as
\[\hat{N}_{i} = \sum_{t}[t|\mu_{\hat{N}_{i}}(t)], \hat{N}_{j} = \sum_{s}[s|\mu_{\hat{N}_{j}}(s)],\]
so, in condition of a whole fuzziness, a cardinal semantic operator performs the following operations:
\begin{enumerate}[label = (\roman*)]
	\item Calculation of partial carries. \\$\hat{p}_{i} = \big \lfloor \frac{\hat{N}_{i}}{n_{i}} \big \rfloor  = \sum\limits_{\big \lfloor  \nicefrac{t}{n_{i}} \big \rfloor  }[\big \lfloor \frac{t}{n_{i}} \big \rfloor |\mu_{\hat{p}_{i}}(\big \lfloor \frac{r}{n_{i}} \big \rfloor )] = \sum\limits_{g}[g|\mu_{\hat{p}_{i}}(g)] ,$\\
	$\hat{p}_{j} = \big \lfloor \frac{\hat{N}_{j}}{n_{j}} \big \rfloor  = \sum\limits_{\big \lfloor \nicefrac{s}{n_{j}} \big \rfloor }[\big \lfloor \frac{s}{n_{j}} \big \rfloor |\mu_{\hat{p}_{j}}\big \lfloor \frac{s}{n_{j}} \big\rfloor ] = \sum\limits_{u}[u|\mu_{\hat{p}_{j}}(u)]$
	\item $\hat{p} = form(\hat{p}_{i}, \hat{p}_{j}) - $formation of a common carry;
	\item The process of calculation an image cardinal and a remainder is analogous to those, that are described in section 3.
\end{enumerate}

In case when $\hat{p}_{i}$ and $\hat{p}_{j}$ are monotonic, the mode of the common carry is defined as
\begin{equation}
	(\hat{p}|1) = min[(\hat{p}_{i}|1), (\hat{p}_{j}|1)].
\end{equation}
If the supports of the partial carries ($g$ and $u$) intersect, a calculation is performed on their union ($\varphi = g \cup u$), so that
	\begin{equation}
	\begin{cases}
		mode(\hat{p}) = min[(\hat{p}_{i}|1), (\hat{p}_{j}|1)] = \hat{p}(\varphi_{0}) = \varphi_{0}; \\
		\mu_{\hat{p}}(\varphi) = max[\mu_{\hat{p}_{i}}(g), \mu_{\hat{p}_{j}}(u)], &  \forall\varphi<\varphi_{0} ;\\
		\mu_{\hat{p}}(\varphi) = min[\mu_{\hat{p}_{i}}(g), \mu_{\hat{p}_{j}}(u)], &  \forall\varphi>\varphi_{0}.\\
	\end{cases} 
\end{equation}
Formation of the fuzzy common carry from partial fuzzy common carries with discrete supports are illustrated in the fig. 6, where $\hat{p}_{i}, \hat{p}_{j}$ are partial carries and $\hat{p}_{o}$ is the fuzzy common carry.
\begin{figure}[h]
	\centering
	\includegraphics[scale=0.35]{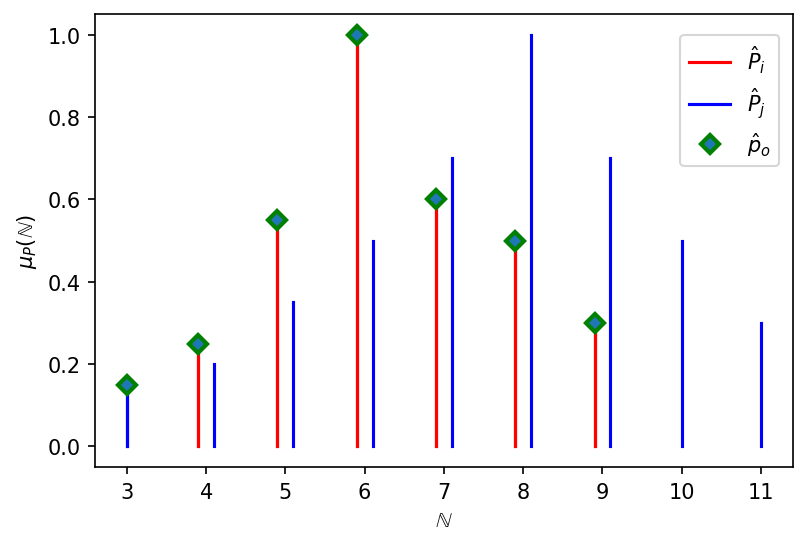}
	\caption{Formation of a discrete fuzzy common carry.}
	\label{fig:fig1}
\end{figure}

\section{Conclusion}
Quite unexpectedly, in fuzzy cardinal semantic operators the common carry is not  chosen, but formed.  This result has, in authors' opinion, an important methodological meaning. In case of superposition of uncertainty of some entities (their intersection or union) the resulting uncertainty may not be a union or intersection of uncertainty of initial entities, but is formed in accordance with the "sense" of the performing procedures. Is that statement reasonable for semantic numeration systems in case of probabilistic type of uncertainty? Does this statement has system-inherent nature?\\
\\
\textbf{Conflicts of Interest:} The authors declare no conflict of interest.
\bibliographystyle{unsrtnat}
\bibliography{references}  





\newpage
\section*{Appendix}

\textbf{Fuzzy L-operator}
\begin{enumerate}
	\item \textbf{( $\tilde{N}_{i}; n_{i}; r_{ij}$)} - fuzzy cardinal
	\paragraph{Carry:} $\\ \\ \tilde{p}_{i} = \big \lfloor \frac{\tilde{N}_{i}}{n_{i}}\big \rfloor  = (\big \lfloor \frac{a_{\tilde{N}_{i}}}{n_{i}}\big \rfloor ; \big \lfloor \frac{m_{\tilde{N}_{i}}}{n_{i}}\big \rfloor ; \big \lfloor \frac{b_{\tilde{N}_{i}}}{n_{i}}\big \rfloor )$
	\paragraph{Remainder:} $\\ \\ \tilde{N}_{i}^{'} = \tilde{N}_{i} - \tilde{p}_{i}n_{i} = (a_{\tilde{N}_{i}}; m_{\tilde{N}_{i}}; b_{\tilde{N}_{i}}) - (a_{\tilde{p}_{i}}n_{i}; m_{\tilde{p}_{i}}n_{i}; b_{\tilde{p}_{i}}n_{i}) =(a_{\tilde{N}_{i}} - b_{\tilde{p}_{i}}n_{i}; m_{\tilde{N}_{i}} - m_{\tilde{p}_{i}}n_{i}; b_{\tilde{N}_{i}} - a_{\tilde{p}_{i}}n_{i})$
	\paragraph{Transformant:} $\\ \\ \tilde{q}_{j} = \tilde{p}_{i}r_{ij} = (a_{\tilde{p}_{i}}r_{ij}; m_{\tilde{p}_{i}}r_{ij}; b_{\tilde{p}_{i}}r_{ij}) = (\big \lfloor \frac{a_{\tilde{N}_{i}}}{n_{i}}\big \rfloor r_{ij}; \big \lfloor \frac{m_{\tilde{N}_{i}}}{n_{i}}\big \rfloor r_{ij}; \big \lfloor \frac{b_{\tilde{N}_{i}}}{n_{i}}\big \rfloor r_{ij})$
	\paragraph{Image cardinal:} $\\ \\ \tilde{N}_{j}^{'} = N_{j} + \tilde{q}_{j}  = N_{j} + \tilde{p}_{i}r_{ij} = N_{j} + \big \lfloor \frac{\tilde{N}_{i}}{n_{i}}\big \rfloor r_{ij} = (N_{j} + \big \lfloor \frac{a_{\tilde{N}_{i}}}{n_{i}}\big \rfloor r_{ij}; N_{j} + \big \lfloor \frac{m_{\tilde{N}_{i}}}{n_{i}}\big \rfloor r_{ij}; N_{j} + \big \lfloor \frac{b_{\tilde{N}_{i}}}{n_{i}}\big \rfloor r_{ij})$
	\item \textbf{($N_{i}; \tilde{n}_{i}; r_{ij}$)} - fuzzy radix
	\paragraph{Carry} $\\ \\ \tilde{p}_{i} = \big \lfloor \frac{N_{i}}{\tilde{n}_{i}}\big \rfloor  = (\big \lfloor \frac{N_{i}}{b_{\tilde{n}_{i}}}\big \rfloor ; \big \lfloor \frac{N_{i}}{m_{\tilde{n}_{i}}}\big \rfloor ; \big \lfloor \frac{N_{i}}{a_{\tilde{n}_{i}}}\big \rfloor )$
	\paragraph{Remainder:} $\\ \\ \tilde{N}_{i}^{'} = N_{i} - \tilde{p}_{i}\tilde{n}_{i} = N_{i} - (a_{\tilde{p}_{i}}a_{\tilde{n}_{i}}; m_{\tilde{p}_{i}}m_{\tilde{n}_{i}}; b_{\tilde{p}_{i}}b_{\tilde{n}_{i}}) = (N_{i} -  b_{\tilde{p}_{i}}b_{\tilde{n}_{i}}; N_{i} - m_{\tilde{p}_{i}}m_{\tilde{n}_{i}};N_{i} - \\ a_{\tilde{p}_{i}}a_{\tilde{n}_{i}})$
	\paragraph{Transformant:} $\\ \\ \tilde{q}_{j} = \tilde{p}_{i}r_{ij} = (a_{\tilde{p}_{i}}r_{ij}; m_{\tilde{p}_{i}}r_{ij}; b_{\tilde{p}_{i}}r_{ij})$
	\paragraph{Image cardinal:} $\\ \\ N_{j}^{'} = N_{j} + \tilde{q}_{j} = N_{j} + \big \lfloor \frac{N_{i}}{\tilde{n}_{i}}\big \rfloor r_{ij} = (N_{j} + \big \lfloor \frac{N_{i}}{a_{\tilde{n}_{i}}}\big \rfloor r_{ij}; N_{j} + \big \lfloor \frac{N_{i}}{m_{\tilde{n}_{i}}}\big \rfloor r_{ij}; N_{j} + \big \lfloor \frac{N_{i}}{b_{\tilde{n}_{i}}}\big \rfloor r_{ij})$
	\item \textbf{($N_{i}; n_{i}; \tilde{r}_{ij}$)} - fuzzy conversion rate
	\paragraph{Carry:} $\\ \\ p_{i} = \big \lfloor \frac{N_{i}}{n_{i}}\big \rfloor $
	\paragraph{Remainder:} $\\ \\ N_{i}^{'} = N_{i} - p_{i}n_{i}$
	\paragraph{Transformant:} $\\ \\ \tilde{q}_{j} = p_{i}\tilde{r}_{ij} = (p_{i}a_{\tilde{r}_{ij}}; p_{i}m_{\tilde{r}_{ij}}; p_{i}b_{\tilde{r}_{ij}})$
	\paragraph{Image cardinal:} $\\ \\ \tilde{N}_{j}^{'} = N_{j} + \tilde{q}_{j} = N_{j} + \big \lfloor \frac{N_{i}}{n_{i}}\big \rfloor \tilde{r}_{ij} = (N_{j} + \big \lfloor \frac{N_{i}}{n_{i}}\big \rfloor a_{\tilde{r}_{ij}}; N_{j} + \big \lfloor \frac{N_{i}}{n_{i}}\big \rfloor m_{\tilde{r}_{ij}}; N_{j} + \big \lfloor \frac{N_{i}}{n_{i}}\big \rfloor b_{\tilde{r}_{ij}})$ 
	\item \textbf{($N_{i}; \tilde{n}_{i}; \tilde{r}_{ij}$)} - fuzzy radix and conversion rate
	\paragraph{Carry:} $\\ \\ \tilde{p}_{i} = \big \lfloor \frac{N_{i}}{\tilde{n}_{i}}\big \rfloor  = (\big \lfloor \frac{N_{i}}{b_{\tilde{n}_{i}}}\big \rfloor ; \big \lfloor \frac{N_{i}}{m_{\tilde{n}_{i}}}\big \rfloor ; \big \lfloor \frac{N_{i}}{a_{\tilde{n}_{i}}}\big \rfloor )$
	\paragraph{Remainder:} $\\ \\ \tilde{N}_{i}^{'} = N_{i} - \tilde{p}_{i}\tilde{n}_{i} = N_{i} - (a_{\tilde{p}_{i}}a_{\tilde{n}_{i}}; m_{\tilde{p}_{i}}m_{\tilde{n}_{i}}; b_{\tilde{p}_{i}}b_{\tilde{n}_{i}}) = (N_{i} -  b_{\tilde{p}_{i}}b_{\tilde{n}_{i}}; N_{i} - m_{\tilde{p}_{i}}m_{\tilde{n}_{i}};N_{i} - a_{\tilde{p}_{i}}a_{\tilde{n}_{i}})$
	\paragraph{Transformant:} $\\ \\ \tilde{q}_{j} = \tilde{p}_{i}\tilde{r}_{ij} = (a_{\tilde{p}_{i}}a_{\tilde{r}_{ij}}; m_{\tilde{p}_{i}}m_{\tilde{r}_{ij}};  b_{\tilde{p}_{i}}b_{\tilde{r}_{ij}})$
	\paragraph{Image cardinal:} $\\ \\ \tilde{N}_{j}^{'} = N_{j} + \tilde{q}_{j} = N_{j} + \big \lfloor \frac{N_{i}}{\tilde{n}_{i}}\big \rfloor \tilde{r}_{ij} = (N_{j} + \big \lfloor \frac{N_{i}}{a_{\tilde{n}_{i}}}\big \rfloor a_{\tilde{r}_{ij}}; N_{j} + \big \lfloor \frac{N_{i}}{m_{\tilde{n}_{i}}}\big \rfloor m_{\tilde{r}_{ij}}; N_{j} + \big \lfloor \frac{N_{i}}{b_{\tilde{n}_{i}}}\big \rfloor b_{\tilde{r}_{ij}})$
	\item \textbf{($\tilde{N}_{i};$ $\tilde{n}_{i};$ $\tilde{r}_{ij}$)} - whole fuzziness
	\paragraph{Carry:} $\\ \\ \tilde{p}_{i} = \big \lfloor \frac{\tilde{N}_{i}}{\tilde{n}_{i}}\big \rfloor  = (\big \lfloor \frac{a_{\tilde{N}_{i}}}{b_{\tilde{n}_{i}}}\big \rfloor ; \big \lfloor \frac{m_{\tilde{N}_{i}}}{m_{\tilde{n}_{i}}}\big \rfloor ; \big \lfloor \frac{b_{\tilde{N}_{i}}}{a_{\tilde{n}_{i}}}\big \rfloor )$
	\paragraph{Remainder:} $\\ \\ \tilde{N}_{i}^{'} = \tilde{N}_{i} - \tilde{p}_{i}\tilde{n}_{i} = (a_{\tilde{N}_{i}}; m_{\tilde{N}_{i}}; b_{\tilde{N}_{i}}) - (a_{\tilde{p}_{i}}a_{\tilde{n}_{i}}; m_{\tilde{p}_{i}}m_{\tilde{n}_{i}}; b_{\tilde{p}_{i}}b_{\tilde{n}_{i}}) = (a_{\tilde{N}_{i}} - b_{\tilde{p}_{i}}b_{\tilde{n}_{i}}; m_{\tilde{N}_{i}} - m_{\tilde{p}_{i}}m_{\tilde{n}_{i}}; b_{\tilde{N}_{i}} - a_{\tilde{p}_{i}}a_{\tilde{n}_{i}})$
	\paragraph{Transformant:} $\\ \\ \tilde{q}_{j} = \tilde{p}_{i}\tilde{r}_{ij} = (a_{\tilde{p}_{i}}a_{\tilde{r}_{ij}}; m_{\tilde{p}_{i}}m_{\tilde{r}_{ij}};  b_{\tilde{p}_{i}}b_{\tilde{r}_{ij}})$
	\paragraph{Image cardinal:} $\\ \\ \tilde{N}_{j}^{'} = \tilde{N}_{j} + \tilde{q}_{j} = N_{j} + \big \lfloor \frac{\tilde{N}_{i}}{\tilde{n}_{i}}\big \rfloor \tilde{r}_{ij} = (N_{j} + \big \lfloor \frac{a_{\tilde{N}_{i}}}{b_{\tilde{n}_{i}}}\big \rfloor a_{\tilde{r}_{ij}}; N_{j} + \big \lfloor \frac{m_{\tilde{N}_{i}}}{m_{\tilde{n}_{i}}}\big \rfloor m_{\tilde{r}_{ij}}; N_{j} + \big \lfloor \frac{b_{\tilde{N}_{i}}}{a_{\tilde{n}_{i}}}\big \rfloor b_{\tilde{r}_{ij}})$
\end{enumerate}

\textbf{Fuzzy \boldmath$\leftindex_{W}{F}$ - operator}
\begin{enumerate}
	\item \textbf{($\tilde{N_{i}},...,\tilde{N_{j}};n_{i},...,n_{j};r_{k}$)} - fuzzy inputs
	\paragraph{Partial carries:}  $\\ \\ \tilde{p}_{i} = \big \lfloor \frac{\tilde{N}_{i}}{n_{i}}\big \rfloor  = (\big \lfloor \frac{a_{\tilde{N}_{i}}}{n_{i}}\big \rfloor ;\big \lfloor \frac{m_{\tilde{N}_{i}}}{n_{i}}\big \rfloor ;\big \lfloor \frac{b_{\tilde{N}_{i}}}{n_{i}}\big \rfloor ) \\$
	$\dots$
	$\\ \tilde{p}_{j} = \big \lfloor \frac{\tilde{N}_{j}}{n_{j}}\big \rfloor  = (\big \lfloor \frac{a_{\tilde{N}_{j}}}{n_{j}}\big \rfloor ;\big \lfloor \frac{m_{\tilde{N}_{j}}}{n_{j}}\big \rfloor ;\big \lfloor \frac{b_{\tilde{N}_{j}}}{n_{j}}\big \rfloor )$
	\paragraph{Fuzzy common carry:} $\\ \\ \tilde{p} = (\text{min}(a_{\tilde{p}_{i}},...,a_{\tilde{p}_{j}}); \text{min}(m_{\tilde{p}_{i}},...,m_{\tilde{p}_{j}}); \text{min}(b_{\tilde{p}_{i}},...,b_{\tilde{p}_{j}})) = \\ = (\text{min}(\big \lfloor \frac{a_{\tilde{N}_{i}}}{n_{i}}\big \rfloor ,...,\big \lfloor \frac{a_{\tilde{N}_{j}}}{n_{j}}\big \rfloor );\text{min}(\big \lfloor \frac{m_{\tilde{N}_{i}}}{n_{i}}\big \rfloor ,...,\big \lfloor \frac{m_{\tilde{N}_{j}}}{n_{j}}\big \rfloor ); \text{min}(\big \lfloor \frac{b_{\tilde{N}_{i}}}{n_{i}}\big \rfloor ,...,\big \lfloor \frac{b_{\tilde{N}_{j}}}{n_{j}}\big \rfloor ))$
	\paragraph{Remainders:} $\\ \\ \tilde{N}_{i}^{'} = \tilde{N}_{i} - \tilde{p}n_{i} = (a_{\tilde{N}_{i}}; m_{\tilde{N}_{i}}; b_{\tilde{N}_{i}}) - (a_{\tilde{p}}n_{i}; m_{\tilde{p}}n_{i}; b_{\tilde{p}}n_{i}) = (a_{\tilde{N}_{i}} - b_{\tilde{p}}n_{i}; m_{\tilde{N}_{i}} - m_{\tilde{p}}n_{i}; b_{\tilde{N}_{i}} - a_{\tilde{p}}n_{i}) = \\ = (a_{\tilde{N}_{i}} - \text{min}(\big \lfloor \frac{b_{\tilde{N}_{i}}}{n_{i}}\big \rfloor ,...,\big \lfloor \frac{b_{\tilde{N}_{j}}}{n_{j}}\big \rfloor )n_{i}; m_{\tilde{N}_{i}} - \text{min}(\big \lfloor \frac{m_{\tilde{N}_{i}}}{n_{i}}\big \rfloor ,...,\big \lfloor \frac{m_{\tilde{N}_{j}}}{n_{j}}\big \rfloor )n_{i}; b_{\tilde{N}_{i}} - \text{min}(\big \lfloor \frac{a_{\tilde{N}_{i}}}{n_{i}}\big \rfloor ,...,\big \lfloor \frac{a_{\tilde{N}_{j}}}{n_{j}}\big \rfloor )n_{i})$\\
	$\dots$\\
	$\\ \tilde{N}_{j}^{'} = \tilde{N}_{j} - \tilde{p}n_{j} = (a_{\tilde{N}_{j}}; m_{\tilde{N}_{j}}; b_{\tilde{N}_{j}}) - (a_{\tilde{p}}n_{j}; m_{\tilde{p}}n_{j}; b_{\tilde{p}}n_{j}) = (a_{\tilde{N}_{j}} - b_{\tilde{p}}n_{j}; m_{\tilde{N}_{j}} - m_{\tilde{p}}n_{j}; b_{\tilde{N}_{j}} - a_{\tilde{p}}n_{j}) =\\= (a_{\tilde{N}_{j}} - \text{min}(\big \lfloor \frac{b_{\tilde{N}_{i}}}{n_{i}}\big \rfloor ,...,\big \lfloor \frac{b_{\tilde{N}_{j}}}{n_{j}}\big \rfloor )n_{i}; m_{\tilde{N}_{j}} - \text{min}(\big \lfloor \frac{m_{\tilde{N}_{i}}}{n_{i}}\big \rfloor ,...,\big \lfloor \frac{m_{\tilde{N}_{j}}}{n_{j}}\big \rfloor )n_{i}; b_{\tilde{N}_{j}} - \text{min}(\big \lfloor \frac{a_{\tilde{N}_{i}}}{n_{i}}\big \rfloor ,...,\big \lfloor \frac{a_{\tilde{N}_{j}}}{n_{j}}\big \rfloor )n_{i})$
	\paragraph{Transformant:} $\\ \\ \tilde{q}_{k} = \tilde{p}r_{k} = (a_{\tilde{p}}r_{k}; m_{\tilde{p}}r_{k}; b_{\tilde{p}}r_{k}) = (\text{min}(\big \lfloor \frac{a_{\tilde{N}_{i}}}{n_{i}}\big \rfloor ,...,\big \lfloor \frac{a_{\tilde{N}_{j}}}{n_{j}}\big \rfloor )r_{k}; \text{min}(\big \lfloor \frac{m_{\tilde{N}_{i}}}{n_{i}}\big \rfloor ,...,\big \lfloor \frac{m_{\tilde{N}_{j}}}{n_{j}}\big \rfloor )r_{k}; \text{min}(\big \lfloor \frac{b_{\tilde{N}_{i}}}{n_{i}}\big \rfloor ,...,\big \lfloor \frac{b_{\tilde{N}_{j}}}{n_{j}}\big \rfloor )r_{k})$
	\paragraph{Image cardinal:} $\\ \\ \tilde{N}_{k}^{'} = N_{k} + \tilde{q}_{k} = N_{k} + \tilde{p}r_{k} = (a_{\tilde{N}_{k}} + a_{\tilde{p}}r_{k}; m_{N_{k}} + m_{\tilde{p}}r_{k}; b_{N_{k}} + b_{\tilde{p}}r_{k}) = (a_{\tilde{N}_{k}} + \text{min}(\big \lfloor \frac{a_{\tilde{N}_{i}}}{n_{i}}\big \rfloor ,...,\big \lfloor \frac{a_{\tilde{N}_{j}}}{n_{j}}\big \rfloor )r_{k}; m_{\tilde{N}_{k}} + \text{min}(\big \lfloor \frac{m_{\tilde{N}_{i}}}{n_{i}}\big \rfloor ,...,\big \lfloor \frac{m_{\tilde{N}_{j}}}{n_{j}}\big \rfloor )r_{k}; b_{\tilde{N}_{k}} + \text{min}(\big \lfloor \frac{b_{\tilde{N}_{i}}}{n_{i}}\big \rfloor ,...,\big \lfloor \frac{b_{\tilde{N}_{j}}}{n_{j}}\big \rfloor )r_{k}))$
	\item \textbf{($N_{i},...,N_{j}; \tilde{n}_{i},...,\tilde{n}_{j}; r_{k}$)} - fuzzy radix
	\paragraph{Partial carries:} $\\ \\ \tilde{p}_{i} = (\big \lfloor \frac{N_{i}}{b_{\tilde{n}_{i}}}\big \rfloor ; \big \lfloor \frac{N_{i}}{m_{\tilde{n}_{i}}}\big \rfloor ; \big \lfloor \frac{N_{i}}{a_{\tilde{n}_{i}}}\big \rfloor )\\$
	$\dots$
	$\\ \tilde{p}_{j} = (\big \lfloor \frac{N_{j}}{b_{\tilde{n}_{j}}}\big \rfloor ; \big \lfloor \frac{N_{j}}{m_{\tilde{n}_{j}}}\big \rfloor ; \big \lfloor \frac{N_{j}}{a_{\tilde{n}_{j}}}\big \rfloor )$
	\paragraph{Fuzzy common carry:} $\\ \\ \tilde{p} = (\text{min}(a_{\tilde{p}_{i}},...,a_{\tilde{p}_{j}}); \text{min}(m_{\tilde{p}_{i}},...,m_{\tilde{p}_{j}}); \text{min}(b_{\tilde{p}_{i}},...b_{\tilde{p}_{j}})) =\\= (\text{min}(\big \lfloor \frac{N_{i}}{b_{\tilde{n}_{i}}}\big \rfloor ,...,\big \lfloor \frac{N_{j}}{b_{\tilde{n}_{j}}}\big \rfloor ); \text{min}(\big \lfloor \frac{N_{i}}{m_{\tilde{n}_{i}}}\big \rfloor ,...,\big \lfloor \frac{N_{j}}{m_{\tilde{n}_{j}}}\big \rfloor ); \text{min}(\big \lfloor \frac{N_{i}}{a_{\tilde{n}_{i}}}\big \rfloor ,...,\big \lfloor \frac{N_{j}}{a_{\tilde{n}_{j}}}\big \rfloor ))$
	\paragraph{Remainders:} $\\ \\ \tilde{N}_{i}^{'} = N_{i} - \tilde{p}\tilde{n}_{i} = N_{i} - (a_{\tilde{n}_{i}}a_{\tilde{p}}; m_{\tilde{n}_{i}}m_{\tilde{p}}; b_{\tilde{n}_{i}}b_{\tilde{p}}) = (N_{i} - b_{\tilde{n}_{i}}b_{\tilde{p}}; N_{i} - m_{\tilde{n}_{i}}m_{\tilde{p}}; N_{i} - a_{\tilde{n}_{i}}a_{\tilde{p}}) =\\= (N_{i} - b_{\tilde{n}_{i}}\text{min}(b_{\tilde{p}_{i}},...b_{\tilde{p}_{j}}); N_{i} - m_{\tilde{n}_{i}}\text{min}(m_{\tilde{p}_{i}},...,m_{\tilde{p}_{j}}); N_{i} - a_{\tilde{n}_{i}}\text{min}(a_{\tilde{p}_{i}},...,a_{\tilde{p}_{j}})\\$
	$\dots$\\
	$\\ \tilde{N}_{j}^{'} = N_{j} - \tilde{p}\tilde{n}_{j} = N_{j} - (a_{\tilde{n}_{j}}a_{\tilde{p}}; m_{\tilde{n}_{j}}m_{\tilde{p}}; b_{\tilde{n}_{j}}b_{\tilde{p}}) = (N_{j} - b_{\tilde{n}_{j}}b_{\tilde{p}}; N_{j} - m_{\tilde{n}_{j}}m_{\tilde{p}}; N_{j} - a_{\tilde{n}_{j}}a_{\tilde{p}}) =\\= (N_{j} - b_{\tilde{n}_{j}}\text{min}(b_{\tilde{p}_{i}},...b_{\tilde{p}_{j}}); N_{j} - m_{\tilde{n}_{j}}\text{min}(m_{\tilde{p}_{i}},...,m_{\tilde{p}_{j}}); N_{j} - a_{\tilde{n}_{j}}\text{min}(a_{\tilde{p}_{i}},...,a_{\tilde{p}_{j}})$
	\paragraph{Transformant:} $\\ \\ \tilde{q}_{k} = \tilde{p}r_{k} = (a_{\tilde{p}}r_{k}; m_{\tilde{p}}r_{k}; b_{\tilde{p}}r_{k}) = (\text{min}(a_{\tilde{p}_{i}},...,a_{\tilde{p}_{j}})r_{k}; \text{min}(m_{\tilde{p}_{i}},...,m_{\tilde{p}_{j}})r_{k}; \text{min}(b_{\tilde{p}_{i}},...b_{\tilde{p}_{j}})r_{k})$
	\paragraph{Image cardinal:} $\\ \\ N_{k}^{'} = N_{k} + \tilde{q}_{k} = N_{k} + \tilde{p}r_{k} = (a_{\tilde{N}_{k}} + a_{\tilde{p}}r_{k}; m_{N_{k}} + m_{\tilde{p}}r_{k}; b_{N_{k}} + b_{\tilde{p}}r_{k}) = (a_{\tilde{N}_{k}} + \text{min}(\big \lfloor \frac{a_{\tilde{N}_{i}}}{n_{i}}\big \rfloor ,...,\big \lfloor \frac{a_{\tilde{N}_{j}}}{n_{j}}\big \rfloor )r_{k}; m_{\tilde{N}_{k}} + \text{min}(\big \lfloor \frac{m_{\tilde{N}_{i}}}{n_{i}}\big \rfloor ,...,\big \lfloor \frac{m_{\tilde{N}_{j}}}{n_{j}}\big \rfloor )r_{k}; b_{\tilde{N}_{k}} + \text{min}(\big \lfloor \frac{b_{\tilde{N}_{i}}}{n_{i}}\big \rfloor ,...,\big \lfloor \frac{b_{\tilde{N}_{j}}}{n_{j}}\big \rfloor )r_{k}))\\$
	\item \textbf{($N_{i},...,N_{j}; n_{i},...,n_{j}; \tilde{r}_{k}$)} - fuzzy conversion rate
	\paragraph{Partial carries:} $\\ \\ p_{i} = \big \lfloor \frac{N_{i}}{n_{i}}\big \rfloor \\$
	$\dots$
	$\\ p_{j} = \big \lfloor \frac{N_{j}}{n_{j}}\big \rfloor \\$
	\paragraph{Fuzzy common carry:} $\\ \\ p = \text{min}(p_{i},...,p_{j}) = \text{min}(\big \lfloor \frac{N_{i}}{n_{i}}\big \rfloor ,...,\big \lfloor \frac{N_{j}}{n_{j}}\big \rfloor )$
	\paragraph{Remainders:} $\\ \\ N_{i}^{'} = N_{i} - pn_{i} = N_{i} - \text{min}(\big \lfloor \frac{N_{i}}{n_{i}}\big \rfloor ,...,\big \lfloor \frac{N_{j}}{n_{j}}\big \rfloor )n_{i}\\$
	$\dots$\\
	$\\ N_{j}^{'} = N_{j} - pn_{j} = N_{j} - \text{min}(\big \lfloor \frac{N_{i}}{n_{i}}\big \rfloor ,...,\big \lfloor \frac{N_{j}}{n_{j}}\big \rfloor )n_{j}$
	\paragraph{Transformant:} $\\ \\ \tilde{q}_{k} = p\tilde{r}_{k} = (pa_{\tilde{r}_{k}}; pm_{\tilde{r}_{k}}; pb_{\tilde{r}_{k}}) = (\text{min}(\big \lfloor \frac{N_{i}}{n_{i}}\big \rfloor ,...,\big \lfloor \frac{N_{j}}{n_{j}}\big \rfloor )a_{\tilde{r}_{k}}; \text{min}(\big \lfloor \frac{N_{i}}{n_{i}}\big \rfloor ,...,\big \lfloor \frac{N_{j}}{n_{j}}\big \rfloor )m_{\tilde{r}_{k}}; \text{min}(\big \lfloor \frac{N_{i}}{n_{i}}\big \rfloor ,...,\big \lfloor \frac{N_{j}}{n_{j}}\big \rfloor )b_{\tilde{r}_{k}})$
	\paragraph{Image cardinal:} $\\ \\ \tilde{N}_{k}^{'} = N_{k} + \tilde{q}_{k} = N_{k} + p\tilde{r}_{k} = (N_{k} + \text{min}(\big \lfloor \frac{N_{i}}{n_{i}}\big \rfloor ,...,\big \lfloor \frac{N_{j}}{n_{j}}\big \rfloor )a_{\tilde{r}_{k}}; N_{k} + \text{min}(\big \lfloor \frac{N_{i}}{n_{i}}\big \rfloor ,...,\big \lfloor \frac{N_{j}}{n_{j}}\big \rfloor )m_{\tilde{r}_{k}}; N_{k} + \text{min}(\big \lfloor \frac{N_{i}}{n_{i}}\big \rfloor ,...,\big \lfloor \frac{N_{j}}{n_{j}}\big \rfloor )b_{\tilde{r}_{k}})$
	\item \textbf{($N_{i},...,N_{j}; \tilde{n}_{i},...,\tilde{n}_{j}; \tilde{r}_{k}$)} - fuzzy radix and conversion rate
	\paragraph{Partial carries:} $\\ \\ \tilde{p}_{i} = \big \lfloor \frac{N_{i}}{\tilde{n}_{i}}\big \rfloor  = (\big \lfloor \frac{N_{i}}{b_{\tilde{n}_{i}}}\big \rfloor ; \big \lfloor \frac{N_{i}}{m_{\tilde{n}_{i}}}\big \rfloor ; \big \lfloor \frac{N_{i}}{a_{\tilde{n}_{i}}}\big \rfloor )\\$
	$\dots$
	$\\ \tilde{p}_{j} = \big \lfloor \frac{N_{j}}{\tilde{n}_{j}}\big \rfloor  = (\big \lfloor \frac{N_{j}}{b_{\tilde{n}_{j}}}\big \rfloor ; \big \lfloor \frac{N_{j}}{m_{\tilde{n}_{j}}}\big \rfloor ; \big \lfloor \frac{N_{j}}{a_{\tilde{n}_{j}}}\big \rfloor )$
	\paragraph{Fuzzy common carry:} $\\ \\ \tilde{p} = (\text{min}(a_{\tilde{p}_{i}},...,a_{\tilde{p}_{j}}); \text{min}(m_{\tilde{p}_{i}},...,m_{\tilde{p}_{j}}); \text{min}(b_{\tilde{p}_{i}},...b_{\tilde{p}_{j}})) = \\ = (\text{min}(\big \lfloor \frac{N_{i}}{b_{\tilde{n}_{i}}}\big \rfloor ,...,\big \lfloor \frac{N_{j}}{b_{\tilde{n}_{j}}}\big \rfloor ); \text{min}(\big \lfloor \frac{N_{i}}{m_{\tilde{n}_{i}}}\big \rfloor ,...,\big \lfloor \frac{N_{i}}{m_{\tilde{n}_{j}}}\big \rfloor ); \text{min}(\big \lfloor \frac{N_{i}}{a_{\tilde{n}_{i}}}\big \rfloor ,...,\big \lfloor \frac{N_{i}}{a_{\tilde{n}_{j}}}\big \rfloor ))$
	\paragraph{Remainders:} $\\ \\ \tilde{N}_{i}^{'} = N_{i} - \tilde{p}\tilde{n}_{i} = N_{i} - (a_{\tilde{p}}a_{\tilde{n}_{i}}; m_{\tilde{p}}m_{\tilde{n}_{i}}; b_{\tilde{p}}b_{\tilde{n}_{i}}) = (N_{i} - b_{\tilde{p}}b_{\tilde{n}_{i}}; N_{i} - m_{\tilde{p}}m_{\tilde{n}_{i}}; N_{i} - a_{\tilde{p}}a_{\tilde{n}_{i}}) = \\ = (N_{i} - \text{min}(b_{\tilde{p}_{i}},...b_{\tilde{p}_{j}})b_{\tilde{n}_{i}}; N_{i} - \text{min}(m_{\tilde{p}_{i}},...,m_{\tilde{p}_{j}})m_{\tilde{n}_{i}}; N_{i} - \text{min}(a_{\tilde{p}_{i}},...,a_{\tilde{p}_{j}})a_{\tilde{n}_{i}}) =\\=(N_{i} - \text{min}(\big \lfloor \frac{N_{i}}{a_{\tilde{n}_{i}}}\big \rfloor ,...,\big \lfloor \frac{N_{i}}{a_{\tilde{n}_{j}}}\big \rfloor )b_{\tilde{n}_{i}}; N_{i} - \text{min}(\big \lfloor \frac{N_{i}}{m_{\tilde{n}_{i}}}\big \rfloor ,...,\big \lfloor \frac{N_{i}}{m_{\tilde{n}_{j}}}\big \rfloor )m_{\tilde{n}_{i}}; N_{i} - \text{min}(\big \lfloor \frac{N_{i}}{b_{\tilde{n}_{i}}}\big \rfloor ,...,\big \lfloor \frac{N_{j}}{b_{\tilde{n}_{j}}}\big \rfloor )a_{\tilde{n}_{i}})\\$
	$\dots$\\
	$\\ \tilde{N}_{j}^{'} = N_{j} - \tilde{p}\tilde{n}_{j} = N_{j} - (a_{\tilde{p}}a_{\tilde{n}_{j}}; m_{\tilde{p}}m_{\tilde{n}_{j}}; b_{\tilde{p}}b_{\tilde{n}_{j}}) = (N_{j} - b_{\tilde{p}}b_{\tilde{n}_{j}}; N_{j} - m_{\tilde{p}}m_{\tilde{n}_{j}}; N_{j} - a_{\tilde{p}}a_{\tilde{n}_{j}}) = \\ = (N_{j} - \text{min}(b_{\tilde{p}_{i}},...b_{\tilde{p}_{j}})b_{\tilde{n}_{j}}; N_{j} - \text{min}(m_{\tilde{p}_{i}},...,m_{\tilde{p}_{j}})m_{\tilde{n}_{j}}; N_{j} - \text{min}(a_{\tilde{p}_{i}},...,a_{\tilde{p}_{j}})a_{\tilde{n}_{j}}) =\\=(N_{j} - \text{min}(\big \lfloor \frac{N_{i}}{a_{\tilde{n}_{i}}}\big \rfloor ,...,\big \lfloor \frac{N_{i}}{a_{\tilde{n}_{j}}}\big \rfloor )b_{\tilde{n}_{j}}; N_{j} - \text{min}(\big \lfloor \frac{N_{i}}{m_{\tilde{n}_{i}}}\big \rfloor ,...,\big \lfloor \frac{N_{i}}{m_{\tilde{n}_{j}}}\big \rfloor )m_{\tilde{n}_{j}}; N_{j} - \text{min}(\big \lfloor \frac{N_{i}}{b_{\tilde{n}_{i}}}\big \rfloor ,...,\big \lfloor \frac{N_{j}}{b_{\tilde{n}_{j}}}\big \rfloor )a_{\tilde{n}_{j}})$
	\paragraph{Transformant:} $\\ \\ \tilde{q}_{k} = \tilde{p}\tilde{r}_{k} = (a_{\tilde{p}}a_{\tilde{r}_{k}}; m_{\tilde{p}}m_{\tilde{r}_{k}}; b_{\tilde{p}}b_{\tilde{r}_{k}}) = (\text{min}(a_{\tilde{p}_{i}},...,a_{\tilde{p}_{j}})a_{\tilde{r}_{k}}; \text{min}(m_{\tilde{p}_{i}},...,m_{\tilde{p}_{j}})m_{\tilde{r}_{k}}; \text{min}(b_{\tilde{p}_{i}},...b_{\tilde{p}_{j}})b_{\tilde{r}_{k}}) =\\= (\text{min}(\big \lfloor \frac{N_{i}}{b_{\tilde{n}_{i}}}\big \rfloor ,...,\big \lfloor \frac{N_{j}}{b_{\tilde{n}_{j}}}\big \rfloor )a_{\tilde{r}_{k}}; \text{min}(\big \lfloor \frac{N_{i}}{m_{\tilde{n}_{i}}}\big \rfloor ,...,\big \lfloor \frac{N_{i}}{m_{\tilde{n}_{j}}}\big \rfloor )m_{\tilde{r}_{k}}; \text{min}(\big \lfloor \frac{N_{i}}{a_{\tilde{n}_{i}}}\big \rfloor ,...,\big \lfloor \frac{N_{i}}{a_{\tilde{n}_{j}}}\big \rfloor )b_{\tilde{r}_{k}})$
	\paragraph{Image cardinal:} $\\ \\ \tilde{N}_{k}^{'} = N_{k} + \tilde{q}_{k} = N_{k} + \tilde{p}\tilde{r}_{k} = (N_{k} + a_{\tilde{p}}a_{\tilde{r}_{k}}; N_{k} + m_{\tilde{p}}m_{\tilde{r}_{k}}; N_{k} + b_{\tilde{p}}b_{\tilde{r}_{k}}) = \\ = (N_{k} + \text{min}(a_{\tilde{p}_{i}},...,a_{\tilde{p}_{j}})a_{\tilde{r}_{k}}; N_{k} + \text{min}(m_{\tilde{p}_{i}},...,m_{\tilde{p}_{j}})m_{\tilde{r}_{k}}; N_{k} +\text{min}(b_{\tilde{p}_{i}},...b_{\tilde{p}_{j}})b_{\tilde{r}_{k}}) = \\ = (N_{k} + \text{min}(\big \lfloor \frac{N_{i}}{b_{\tilde{n}_{i}}}\big \rfloor ,...,\big \lfloor \frac{N_{j}}{b_{\tilde{n}_{j}}}\big \rfloor )a_{\tilde{r}_{k}}; N_{k} + \text{min}(\big \lfloor \frac{N_{i}}{m_{\tilde{n}_{i}}}\big \rfloor ,...,\big \lfloor \frac{N_{i}}{m_{\tilde{n}_{j}}}\big \rfloor )m_{\tilde{r}_{k}}; N_{k} +\text{min}(\big \lfloor \frac{N_{i}}{a_{\tilde{n}_{i}}}\big \rfloor ,...,\big \lfloor \frac{N_{j}}{a_{\tilde{n}_{j}}}\big \rfloor )b_{\tilde{r}_{k}})$
	\item \textbf{($\tilde{N}_{i},...,\tilde{N}_{j}; \tilde{n}_{i},...,\tilde{n}_{j};\tilde{r}_{k}$)} - whole fuzziness
	\paragraph{Partial carries:} $\\ \\ \tilde{p}_{i} = (\big \lfloor \frac{a_{\tilde{N}_{i}}}{b_{\tilde{n}_{i}}}\big \rfloor ; \big \lfloor \frac{m_{\tilde{N}_{i}}}{m_{\tilde{n}_{i}}}\big \rfloor ; \big \lfloor \frac{b_{\tilde{N}_{i}}}{a_{\tilde{n}_{i}}}\big \rfloor )\\$
	$\dots$
	$\\ \tilde{p}_{j} = (\big \lfloor \frac{a_{\tilde{N}_{j}}}{b_{\tilde{n}_{j}}}\big \rfloor ; \big \lfloor \frac{m_{\tilde{N}_{j}}}{m_{\tilde{n}_{j}}}\big \rfloor ; \big \lfloor \frac{b_{\tilde{N}_{j}}}{a_{\tilde{n}_{j}}}\big \rfloor )$
	\paragraph{Fuzzy common carry:} $\\ \\ \tilde{p} = (\text{min}(a_{\tilde{p}_{i}},...,a_{\tilde{p}_{j}}); \text{min}(m_{\tilde{p}_{i}},...,m_{\tilde{p}_{j}}); \text{min}(b_{\tilde{p}_{i}},...b_{\tilde{p}_{j}})) =\\= (\text{min}(\big \lfloor \frac{a_{\tilde{N}_{i}}}{b_{\tilde{n}_{i}}}\big \rfloor ,...,\big \lfloor \frac{a_{\tilde{N}_{j}}}{b_{\tilde{n}_{j}}}\big \rfloor ); \text{min}(\big \lfloor \frac{m_{\tilde{N}_{i}}}{m_{\tilde{n}_{i}}}\big \rfloor ,...,\big \lfloor \frac{m_{\tilde{N}_{j}}}{m_{\tilde{n}_{j}}}\big \rfloor ); \text{min}(\big \lfloor \frac{b_{\tilde{N}_{i}}}{a_{\tilde{n}_{i}}}\big \rfloor ,...,\big \lfloor \frac{b_{\tilde{N}_{j}}}{a_{\tilde{n}_{j}}}\big \rfloor ))$
	\paragraph{Remainders:} $\\ \\ \tilde{N}_{i}^{'} = \tilde{N}_{i} - \tilde{p}\tilde{n}_{i} = (a_{\tilde{N}_{i}}; m_{\tilde{N}_{i}}; b_{\tilde{N}_{i}}) - (a_{\tilde{p}}a_{\tilde{n}_{i}}; m_{\tilde{p}}m_{\tilde{n}_{i}}; b_{\tilde{p}}b_{\tilde{n}_{i}}) = (a_{\tilde{N}_{i}} - b_{\tilde{p}}b_{\tilde{n}_{i}}; m_{\tilde{N}_{i}} - m_{\tilde{p}}m_{\tilde{n}_{i}}; b_{\tilde{N}_{i}} - a_{\tilde{p}}a_{\tilde{n}_{i}}) =\\= (a_{\tilde{N}_{i}} - \text{min}(\big \lfloor \frac{b_{\tilde{N}_{i}}}{a_{\tilde{n}_{i}}}\big \rfloor ,...,\big \lfloor \frac{b_{\tilde{N}_{j}}}{a_{\tilde{n}_{j}}}\big \rfloor )b_{\tilde{n}_{i}}; m_{\tilde{N}_{i}} - \text{min}(\big \lfloor \frac{m_{\tilde{N}_{i}}}{m_{\tilde{n}_{i}}}\big \rfloor ,...,\big \lfloor \frac{m_{\tilde{N}_{j}}}{m_{\tilde{n}_{j}}}\big \rfloor )m_{\tilde{n}_{i}}; b_{\tilde{N}_{i}} - \text{min}(\big \lfloor \frac{a_{\tilde{N}_{i}}}{b_{\tilde{n}_{i}}}\big \rfloor ,...,\big \lfloor \frac{a_{\tilde{N}_{j}}}{b_{\tilde{n}_{j}}}\big \rfloor )a_{\tilde{n}_{i}})\\$
	$\dots$\\
	$\\ \tilde{N}_{j}^{'} = \tilde{N}_{j} - \tilde{p}\tilde{n}_{j} = (a_{\tilde{N}_{j}}; m_{\tilde{N}_{j}}; b_{\tilde{N}_{j}}) - (a_{\tilde{p}}a_{\tilde{n}_{j}}; m_{\tilde{p}}m_{\tilde{n}_{j}}; b_{\tilde{p}}b_{\tilde{n}_{j}}) = (a_{\tilde{N}_{j}} - b_{\tilde{p}}b_{\tilde{n}_{j}}; m_{\tilde{N}_{j}} - m_{\tilde{p}}m_{\tilde{n}_{j}}; b_{\tilde{N}_{j}} - a_{\tilde{p}}a_{\tilde{n}_{j}}) =\\= (a_{\tilde{N}_{j}} - \text{min}(\big \lfloor \frac{b_{\tilde{N}_{i}}}{a_{\tilde{n}_{i}}}\big \rfloor ,...,\big \lfloor \frac{b_{\tilde{N}_{j}}}{a_{\tilde{n}_{j}}}\big \rfloor )b_{\tilde{n}_{j}}; m_{\tilde{N}_{j}} - \text{min}(\big \lfloor \frac{m_{\tilde{N}_{i}}}{m_{\tilde{n}_{i}}}\big \rfloor ,...,\big \lfloor \frac{m_{\tilde{N}_{j}}}{m_{\tilde{n}_{j}}}\big \rfloor )m_{\tilde{n}_{j}}; b_{\tilde{N}_{j}} - \text{min}(\big \lfloor \frac{a_{\tilde{N}_{i}}}{b_{\tilde{n}_{i}}}\big \rfloor ,...,\big \lfloor \frac{a_{\tilde{N}_{j}}}{b_{\tilde{n}_{j}}}\big \rfloor )a_{\tilde{n}_{j}})$
	\paragraph{Transformant:} $\\ \\ \tilde{q}_{k} = \tilde{p}\tilde{r}_{k} = (a_{\tilde{p}}a_{\tilde{r}_{k}}; m_{\tilde{p}}m_{\tilde{r}_{k}}; b_{\tilde{p}}b_{\tilde{r}_{k}}) = (\text{min}(\big \lfloor \frac{a_{\tilde{N}_{i}}}{b_{\tilde{n}_{i}}}\big \rfloor ,...,\big \lfloor \frac{a_{\tilde{N}_{j}}}{b_{\tilde{n}_{j}}}\big \rfloor )a_{\tilde{r}_{k}}; \text{min}(\big \lfloor \frac{m_{\tilde{N}_{i}}}{m_{\tilde{n}_{i}}}\big \rfloor ,...,\big \lfloor \frac{m_{\tilde{N}_{j}}}{m_{\tilde{n}_{j}}}\big \rfloor )m_{\tilde{r}_{k}};\text{min}(\big \lfloor \frac{b_{\tilde{N}_{i}}}{a_{\tilde{n}_{i}}}\big \rfloor ,...,\big \lfloor \frac{b_{\tilde{N}_{j}}}{a_{\tilde{n}_{j}}}\big \rfloor )b_{\tilde{r}_{k}})$
	\paragraph{Image cardinal:} $\\ \\ \tilde{N}_{k}^{'} = N_{k} + \tilde{q}_{k} = \tilde{N}_{k} + \tilde{p}\tilde{r}_{k} = (\tilde{N}_{k} + a_{\tilde{p}}a_{\tilde{r}_{k}}; \tilde{N}_{k} + m_{\tilde{p}}m_{\tilde{r}_{k}}; \tilde{N}_{k} + b_{\tilde{p}}b_{\tilde{r}_{k}}) =\\= (\tilde{N}_{k} + \text{min}(\big \lfloor \frac{a_{\tilde{N}_{i}}}{b_{\tilde{n}_{i}}}\big \rfloor ,...,\big \lfloor \frac{a_{\tilde{N}_{j}}}{b_{\tilde{n}_{j}}}\big \rfloor )a_{\tilde{r}_{k}}; \tilde{N}_{k} + \text{min}(\big \lfloor \frac{m_{\tilde{N}_{i}}}{m_{\tilde{n}_{i}}}\big \rfloor ,...,\big \lfloor \frac{m_{\tilde{N}_{j}}}{m_{\tilde{n}_{j}}}\big \rfloor )m_{\tilde{r}_{k}}; \tilde{N}_{k} + \text{min}(\big \lfloor \frac{b_{\tilde{N}_{i}}}{a_{\tilde{n}_{i}}}\big \rfloor ,...,\big \lfloor \frac{b_{\tilde{N}_{j}}}{a_{\tilde{n}_{j}}}\big \rfloor )b_{\tilde{r}_{k}})$
\end{enumerate}	

\textbf{Fuzzy \boldmath$D_{V}$-operator}
\begin{enumerate}
	\item \textbf{($\tilde{N}_{i}; n_{i}; r_{k},...,r_{l}$)} - fuzzy inputs
	\paragraph{Carry:} $\\ \\ \tilde{p}_{i} = \big \lfloor \frac{\tilde{N}_{i}}{n_{i}}\big \rfloor  = (\big \lfloor \frac{a_{\tilde{N}_{i}}}{n_{i}}\big \rfloor ; \big \lfloor \frac{m_{\tilde{N}_{i}}}{n_{i}}\big \rfloor ; \big \lfloor \frac{b_{\tilde{N}_{i}}}{n_{i}}\big \rfloor )$
	\paragraph{Remainder:} $\\ \\ \tilde{N}_{i}^{'} = \tilde{N}_{i} - \tilde{p}_{i}n_{i} = (a_{\tilde{N}_{i}}; m_{\tilde{N}_{i}}; b_{\tilde{N}_{i}}) - (a_{\tilde{p}_{i}}n_{i}; m_{\tilde{p}_{i}}n_{i}; b_{\tilde{p}_{i}}n_{i}) =(a_{\tilde{N}_{i}} - b_{\tilde{p}_{i}}n_{i}; m_{\tilde{N}_{i}} - m_{\tilde{p}_{i}}n_{i}; b_{\tilde{N}_{i}} - a_{\tilde{p}_{i}}n_{i}) =\\= (a_{\tilde{N}_{i}} - \big \lfloor \frac{b_{\tilde{N}_{i}}}{n_{i}}\big \rfloor n_{i}; m_{\tilde{N}_{i}} - \big \lfloor \frac{m_{\tilde{N}_{i}}}{n_{i}}\big \rfloor n_{i}; b_{\tilde{N}_{i}} - \big \lfloor \frac{a_{\tilde{N}_{i}}}{n_{i}}\big \rfloor n_{i})$
	\paragraph{Transformants:} $\\ \\ \tilde{q}_{k} = \tilde{p}_{i}r_{ik} = \big \lfloor \frac{\tilde{N}_{i}}{n_{i}}\big \rfloor r_{ik} = (\big \lfloor \frac{a_{\tilde{N}_{i}}}{n_{i}}\big \rfloor r_{ik}; \big \lfloor \frac{m_{\tilde{N}_{i}}}{n_{i}}\big \rfloor r_{ik}; \big \lfloor \frac{b_{\tilde{N}_{i}}}{n_{i}}\big \rfloor r_{ik})\\$
	$\dots$
	$\\ \tilde{q}_{l} = \tilde{p}_{i}r_{il} = \big \lfloor \frac{\tilde{N}_{i}}{n_{i}}\big \rfloor r_{il} = (\big \lfloor \frac{a_{\tilde{N}_{i}}}{n_{i}}\big \rfloor r_{il}; \big \lfloor \frac{m_{\tilde{N}_{i}}}{n_{i}}\big \rfloor r_{il}; \big \lfloor \frac{b_{\tilde{N}_{i}}}{n_{i}}\big \rfloor r_{il})$
	\paragraph{Image cardinals:} $\\ \\ \tilde{N}_{k}^{'} = N_{k} + \tilde{q}_{k} = N_{k} + \tilde{p}_{i}r_{k} = N_{k} + \big \lfloor \frac{\tilde{N}_{i}}{n_{i}}\big \rfloor r_{k} = (N_{k} + \big \lfloor \frac{a_{\tilde{N}_{i}}}{n_{i}}\big \rfloor r_{k}; N_{k} + \big \lfloor \frac{m_{\tilde{N}_{i}}}{n_{i}}\big \rfloor r_{k}; N_{k} + \big \lfloor \frac{b_{\tilde{N}_{i}}}{n_{i}}\big \rfloor r_{k})\\$
	$\dots$\\
	$\\ \tilde{N}_{l}^{'} = N_{l} + \tilde{q}_{l} = N_{l} + \tilde{p}_{i}r_{l} = N_{l} + \big \lfloor \frac{\tilde{N}_{i}}{n_{i}}\big \rfloor r_{l} = (N_{l} + \big \lfloor \frac{a_{\tilde{N}_{i}}}{n_{i}}\big \rfloor r_{l}; N_{l} + \big \lfloor \frac{m_{\tilde{N}_{i}}}{n_{i}}\big \rfloor r_{l}; N_{l} + \big \lfloor \frac{b_{\tilde{N}_{i}}}{n_{i}}\big \rfloor r_{l})$
	\item \textbf{($N_{i}; \tilde{n}_{i}; r_{ik},...,r_{il};$)} - fuzzy radix
	\paragraph{Carry:} $\\ \\ \tilde{p}_{i} = \big \lfloor \frac{N_{i}}{\tilde{n}_{i}}\big \rfloor  = (\big \lfloor \frac{N_{i}}{b_{\tilde{n}_{i}}}\big \rfloor ; \big \lfloor \frac{N_{i}}{m_{\tilde{n}_{i}}}\big \rfloor ; \big \lfloor \frac{N_{i}}{a_{\tilde{n}_{i}}}\big \rfloor )$
	\paragraph{Remainder:} $\\ \\ \tilde{N}_{i}^{'} = N_{i} - \tilde{p}_{i}\tilde{n}_{i} = N_{i} - (a_{\tilde{p}_{i}}a_{\tilde{n}_{i}}; m_{\tilde{p}_{i}}m_{\tilde{n}_{i}}; b_{\tilde{p}_{i}}b_{\tilde{n}_{i}}) = (N_{i} -  b_{\tilde{p}_{i}}b_{\tilde{n}_{i}}; N_{i} - m_{\tilde{p}_{i}}m_{\tilde{n}_{i}};N_{i} - \\ a_{\tilde{p}_{i}}a_{\tilde{n}_{i}})$
	\paragraph{Transformants:} $\\ \\ \tilde{q}_{k} = \tilde{p}_{i}r_{ik} = \big \lfloor \frac{N_{i}}{\tilde{n}_{i}}\big \rfloor r_{ik} = (\big \lfloor \frac{N_{i}}{b_{\tilde{n}_{i}}}\big \rfloor r_{ik}; \big \lfloor \frac{N_{i}}{m_{\tilde{n}_{i}}}\big \rfloor r_{ik}; \big \lfloor \frac{N_{i}}{a_{\tilde{n}_{i}}}\big \rfloor r_{ik})\\$
	$\dots$
	$\\ \tilde{q}_{l} = \tilde{p}_{i}r_{il} = \big \lfloor \frac{N_{i}}{\tilde{n}_{i}}\big \rfloor r_{il} = (\big \lfloor \frac{N_{i}}{b_{\tilde{n}_{i}}}\big \rfloor r_{il}; \big \lfloor \frac{N_{i}}{m_{\tilde{n}_{i}}}\big \rfloor r_{il}; \big \lfloor \frac{N_{i}}{a_{\tilde{n}_{i}}}\big \rfloor r_{il})$
	\paragraph{Image cardinals:} $\\ \\ \tilde{N}_{k}^{'} = N_{k} + \tilde{q}_{k} = N_{k} + \tilde{p}_{i}r_{ik} = N_{k} + \big \lfloor \frac{N_{i}}{\tilde{n}_{i}}\big \rfloor r_{ik} = (N_{k} + \big \lfloor \frac{N_{i}}{b_{\tilde{n}_{i}}}\big \rfloor r_{ik}; N_{k} + \big \lfloor \frac{N_{i}}{m_{\tilde{n}_{i}}}\big \rfloor r_{ik}; N_{k} + \big \lfloor \frac{N_{i}}{a_{\tilde{n}_{i}}}\big \rfloor r_{ik})\\$
	$\dots$\\
	$\\ \tilde{N}_{l}^{'} = N_{l} + \tilde{q}_{l} = N_{l} + \tilde{p}_{i}r_{il} = N_{l} + \big \lfloor \frac{N_{i}}{\tilde{n}_{i}}\big \rfloor r_{il} = (N_{l} + \big \lfloor \frac{N_{i}}{b_{\tilde{n}_{i}}}\big \rfloor r_{il}; N_{l} + \big \lfloor \frac{N_{i}}{m_{\tilde{n}_{i}}}\big \rfloor r_{il}; N_{l} + \big \lfloor \frac{N_{i}}{a_{\tilde{n}_{i}}}\big \rfloor r_{il})$
	\item \textbf{($N_{i}; n_{i}; \tilde{r}_{ik},...\tilde{r}_{il}$)} - fuzzy conversion rate
	\paragraph{Carry:} $\\ \\ p_{i} = \big \lfloor \frac{N_{i}}{n_{i}}\big \rfloor $
	\paragraph{Remainder:} $\\ \\ N_{i}^{'} = N_{i} - p_{i}n_{i}$
	\paragraph{Transformants:} $\\ \\ \tilde{q}_{k} = p_{i}\tilde{r}_{ik} = (p_{i}a_{\tilde{r}_{ik}}; p_{i}m_{\tilde{r}_{ik}}; p_{i}b_{\tilde{r}_{ik}}) = \big \lfloor \frac{N_{i}}{n_{i}}\big \rfloor a_{\tilde{r}_{ik}}; \big \lfloor \frac{N_{i}}{n_{i}}\big \rfloor m_{\tilde{r}_{ik}}; \big \lfloor \frac{N_{i}}{n_{i}}\big \rfloor b_{\tilde{r}_{ik}})\\$
	$\dots$
	$\\ \tilde{q}_{l} = p_{i}\tilde{r}_{il} = (p_{i}a_{\tilde{r}_{il}}; p_{i}m_{\tilde{r}_{il}}; p_{i}b_{\tilde{r}_{il}}) = \big \lfloor \frac{N_{i}}{n_{i}}\big \rfloor a_{\tilde{r}_{il}}; \big \lfloor \frac{N_{i}}{n_{i}}\big \rfloor m_{\tilde{r}_{il}}; \big \lfloor \frac{N_{i}}{n_{i}}\big \rfloor b_{\tilde{r}_{il}})$
	\paragraph{Image cardinals:} $\\ \\ \tilde{N}_{k}^{'} = N_{k} + \tilde{q}_{k} = N_{k} + p_{i}\tilde{r}_{ik} = (N_{k} + p_{i}a_{\tilde{r}_{ik}}; N_{k} + p_{i}m_{\tilde{r}_{ik}}; N_{k} + p_{i}b_{\tilde{r}_{ik}}) =\\= (N_{k} + \big \lfloor \frac{N_{i}}{n_{i}}\big \rfloor a_{\tilde{r}_{ik}}; N_{k} + \big \lfloor \frac{N_{i}}{n_{i}}\big \rfloor m_{\tilde{r}_{ik}}; N_{k} + \big \lfloor \frac{N_{i}}{n_{i}}\big \rfloor b_{\tilde{r}_{ik}})\\$
	$\dots$\\
	$\\ \tilde{N}_{l}^{'} = N_{l} + \tilde{q}_{l} = N_{l} + p_{i}\tilde{r}_{il} = (N_{l} + p_{i}a_{\tilde{r}_{il}}; N_{l} + p_{i}m_{\tilde{r}_{il}}; N_{l} + p_{i}b_{\tilde{r}_{il}}) =\\= (N_{l} + \big \lfloor \frac{N_{i}}{n_{i}}\big \rfloor a_{\tilde{r}_{il}}; N_{l} + \big \lfloor \frac{N_{i}}{n_{i}}\big \rfloor m_{\tilde{r}_{il}}; N_{l} + \big \lfloor \frac{N_{i}}{n_{i}}\big \rfloor b_{\tilde{r}_{il}})$
	\item \textbf{($N_{i}; \tilde{n}_{i}; \tilde{r}_{ik},...,\tilde{r}_{il}$)} - fuzzy Partial carries and conversion rate
	\paragraph{Carry:} $\\ \\ \tilde{p}_{i} = \big \lfloor \frac{N_{i}}{\tilde{n}_{i}}\big \rfloor  = \big \lfloor \frac{N_{i}}{b_{\tilde{n}_{i}}}\big \rfloor ; \big \lfloor \frac{N_{i}}{m_{\tilde{n}_{i}}}\big \rfloor ; \big \lfloor \frac{N_{i}}{a_{\tilde{n}_{i}}}\big \rfloor $
	\paragraph{Remainder:} $\\ \\ \tilde{N}_{i}^{'} = N_{i} - \tilde{p}_{i}\tilde{n}_{i} = N_{i} - (a_{\tilde{p}_{i}}a_{\tilde{n}_{i}}; m_{\tilde{p}_{i}}m_{\tilde{n}_{i}}; b_{\tilde{p}_{i}}b_{\tilde{n}_{i}}) = (N_{i} -  b_{\tilde{p}_{i}}b_{\tilde{n}_{i}}; N_{i} - m_{\tilde{p}_{i}}m_{\tilde{n}_{i}};N_{i} - \\ a_{\tilde{p}_{i}}a_{\tilde{n}_{i}})$
	\paragraph{Transformants:} $\\ \\ \tilde{q}_{k} = \tilde{p}_{i}\tilde{r}_{ik} = (a_{\tilde{p}_{i}}a_{\tilde{r}_{ik}}; m_{\tilde{p}_{i}}m_{\tilde{r}_{ik}}; b_{\tilde{p}_{i}}b_{\tilde{r}_{ik}}) = (\big \lfloor \frac{N_{i}}{b_{\tilde{n}_{i}}}\big \rfloor a_{\tilde{r}_{ik}}; \big \lfloor \frac{N_{i}}{m_{\tilde{n}_{i}}}\big \rfloor m_{\tilde{r}_{ik}}; \big \lfloor \frac{N_{i}}{a_{\tilde{n}_{i}}}\big \rfloor b_{\tilde{r}_{ik}})\\$
	$\dots$
	$\\ \tilde{q}_{l} = \tilde{p}_{i}\tilde{r}_{il} = (a_{\tilde{p}_{i}}a_{\tilde{r}_{il}}; m_{\tilde{p}_{i}}m_{\tilde{r}_{il}}; b_{\tilde{p}_{i}}b_{\tilde{r}_{il}}) = (\big \lfloor \frac{N_{i}}{b_{\tilde{n}_{i}}}\big \rfloor a_{\tilde{r}_{il}}; \big \lfloor \frac{N_{i}}{m_{\tilde{n}_{i}}}\big \rfloor m_{\tilde{r}_{il}}; \big \lfloor \frac{N_{i}}{a_{\tilde{n}_{i}}}\big \rfloor b_{\tilde{r}_{il}})$
	\paragraph{Image cardinals:} $\\ \\ \tilde{N}_{k}^{'} = N_{k} + \tilde{q}_{k} = N_{k} + \tilde{p}_{i}\tilde{r}_{ik} = (N_{k} + a_{\tilde{p}_{i}}a_{\tilde{r}_{ik}}; N_{k} + m_{\tilde{p}_{i}}m_{\tilde{r}_{ik}}; N_{k} + b_{\tilde{p}_{i}}b_{\tilde{r}_{ik}}) =\\= (N_{k} + \big \lfloor \frac{N_{i}}{b_{\tilde{n}_{i}}}\big \rfloor a_{\tilde{r}_{ik}}; N_{k} + \big \lfloor \frac{N_{i}}{m_{\tilde{n}_{i}}}\big \rfloor m_{\tilde{r}_{ik}}; N_{k} + \big \lfloor \frac{N_{i}}{a_{\tilde{n}_{i}}}\big \rfloor b_{\tilde{r}_{ik}})\\$
	$\dots$\\
	$\\ \tilde{N}_{l}^{'} = N_{l} + \tilde{q}_{l} = N_{l} + \tilde{p}_{i}\tilde{r}_{il} = (N_{l} + a_{\tilde{p}_{i}}a_{\tilde{r}_{il}}; N_{l} + m_{\tilde{p}_{i}}m_{\tilde{r}_{il}}; N_{l} + b_{\tilde{p}_{i}}b_{\tilde{r}_{il}}) =\\= (N_{l} + \big \lfloor \frac{N_{i}}{b_{\tilde{n}_{i}}}\big \rfloor a_{\tilde{r}_{il}}; N_{l} + \big \lfloor \frac{N_{i}}{m_{\tilde{n}_{i}}}\big \rfloor m_{\tilde{r}_{il}}; N_{l} + \big \lfloor \frac{N_{i}}{a_{\tilde{n}_{i}}}\big \rfloor b_{\tilde{r}_{il}})$
	\item \textbf{($\tilde{N}_{i}; \tilde{n}_{i}; \tilde{r}_{ik},...,\tilde{r}_{il}$)} - whole fuzziness
	\paragraph{Carry:} $\\ \\ \tilde{p}_{i} = \big \lfloor \frac{\tilde{N}_{i}}{\tilde{n}_{i}}\big \rfloor  = (\big \lfloor \frac{a_{\tilde{N}_{i}}}{b_{\tilde{n}_{i}}}\big \rfloor ; \big \lfloor \frac{m_{\tilde{N}_{i}}}{m_{\tilde{n}_{i}}}\big \rfloor ; \big \lfloor \frac{b_{\tilde{N}_{i}}}{a_{\tilde{n}_{i}}}\big \rfloor )$
	\paragraph{Remainder:} $\\ \\ \tilde{N}_{i}^{'} = \tilde{N}_{i} - \tilde{p}_{i}\tilde{n}_{i} = (a_{\tilde{N}_{i}}; m_{\tilde{N}_{i}}; b_{\tilde{N}_{i}}) - (a_{\tilde{p}_{i}}a_{\tilde{n}_{i}}; m_{\tilde{p}_{i}}m_{\tilde{n}_{i}}; b_{\tilde{p}_{i}}b_{\tilde{n}_{i}}) = (a_{\tilde{N}_{i}} - b_{\tilde{p}_{i}}b_{\tilde{n}_{i}}; m_{\tilde{N}_{i}} - m_{\tilde{p}_{i}}m_{\tilde{n}_{i}}; b_{\tilde{N}_{i}} - a_{\tilde{p}_{i}}a_{\tilde{n}_{i}})$
	\paragraph{Transformants:} $\\ \\ \tilde{q}_{k} = \tilde{p}_{i}\tilde{r}_{ik} = (a_{\tilde{p}_{i}}a_{\tilde{r}_{ik}}; m_{\tilde{p}_{i}}m_{\tilde{r}_{ik}};  b_{\tilde{p}_{i}}b_{\tilde{r}_{ik}}) = (\big \lfloor \frac{a_{\tilde{N}_{i}}}{b_{\tilde{n}_{i}}}\big \rfloor a_{\tilde{r}_{ik}}; \big \lfloor \frac{m_{\tilde{N}_{i}}}{m_{\tilde{n}_{i}}}\big \rfloor m_{\tilde{r}_{ik}};  \big \lfloor \frac{b_{\tilde{N}_{i}}}{a_{\tilde{n}_{i}}}\big \rfloor b_{\tilde{r}_{ik}})\\$
	$\dots$
	$\\ \tilde{q}_{l} = \tilde{p}_{i}\tilde{r}_{il} = (a_{\tilde{p}_{i}}a_{\tilde{r}_{il}}; m_{\tilde{p}_{i}}m_{\tilde{r}_{il}};  b_{\tilde{p}_{i}}b_{\tilde{r}_{il}}) = (\big \lfloor \frac{a_{\tilde{N}_{i}}}{b_{\tilde{n}_{i}}}\big \rfloor a_{\tilde{r}_{il}}; \big \lfloor \frac{m_{\tilde{N}_{i}}}{m_{\tilde{n}_{i}}}\big \rfloor m_{\tilde{r}_{il}};  \big \lfloor \frac{b_{\tilde{N}_{i}}}{a_{\tilde{n}_{i}}}\big \rfloor b_{\tilde{r}_{il}})$
	\paragraph{Image cardinals:} $\\ \\ \tilde{N}_{k}^{'} = N_{k} + \tilde{q}_{k} = N_{k} + \tilde{p}_{i}\tilde{r}_{ik} =  N_{k} + \big \lfloor \frac{\tilde{N}_{i}}{\tilde{n}_{i}}\big \rfloor \tilde{r}_{ik} = (N_{k} + \big \lfloor \frac{a_{\tilde{N}_{i}}}{b_{\tilde{n}_{i}}}\big \rfloor a_{\tilde{r}_{ik}}; N_{k} + \big \lfloor \frac{m_{\tilde{N}_{i}}}{m_{\tilde{n}_{i}}}\big \rfloor m_{\tilde{r}_{ik}}; N_{k} + \big \lfloor \frac{b_{\tilde{N}_{i}}}{a_{\tilde{n}_{i}}}\big \rfloor b_{\tilde{r}_{ik}})\\$
	$\dots$
	$\\ \tilde{N}_{l}^{'} = N_{l} + \tilde{q}_{l} = N_{l} + \tilde{p}_{i}\tilde{r}_{il} = N_{l} + \big \lfloor \frac{\tilde{N}_{i}}{\tilde{n}_{i}}\big \rfloor \tilde{r}_{il} = (N_{l} + \big \lfloor \frac{a_{\tilde{N}_{i}}}{b_{\tilde{n}_{i}}}\big \rfloor a_{\tilde{r}_{il}}; N_{l} + \big \lfloor \frac{m_{\tilde{N}_{i}}}{m_{\tilde{n}_{i}}}\big \rfloor m_{\tilde{r}_{il}}; N_{l} + \big \lfloor \frac{b_{\tilde{N}_{i}}}{a_{\tilde{n}_{i}}}\big \rfloor b_{\tilde{r}_{il}})$
\end{enumerate}

\textbf{Fuzzy \boldmath$\leftindex_{W}{M}_{V}$-operator}
\begin{enumerate}
	\item \textbf{($\tilde{N}_{i},...,\tilde{N}_{j}; n_{i},...,n_{j}; r_{k},...,r_{l}$)} - fuzzy inputs
	\paragraph{Partial carries:}  $\\ \\ \tilde{p}_{i} = \big \lfloor \frac{\tilde{N}_{i}}{n_{i}}\big \rfloor  = (\big \lfloor \frac{a_{\tilde{N}_{i}}}{n_{i}}\big \rfloor ;\big \lfloor \frac{m_{\tilde{N}_{i}}}{n_{i}}\big \rfloor ;\big \lfloor \frac{b_{\tilde{N}_{i}}}{n_{i}}\big \rfloor ) \\$
	$\dots$
	$\\ \tilde{p}_{j} = \big \lfloor \frac{\tilde{N}_{j}}{n_{j}}\big \rfloor  = (\big \lfloor \frac{a_{\tilde{N}_{j}}}{n_{j}}\big \rfloor ;\big \lfloor \frac{m_{\tilde{N}_{j}}}{n_{j}}\big \rfloor ;\big \lfloor \frac{b_{\tilde{N}_{j}}}{n_{j}}\big \rfloor )$
	\paragraph{Fuzzy common carry:} $\\ \\ \tilde{p} = (\text{min}(a_{\tilde{p}_{i}},...,a_{\tilde{p}_{j}}); \text{min}(m_{\tilde{p}_{i}},...,m_{\tilde{p}_{j}}); \text{min}(b_{\tilde{p}_{i}},...,b_{\tilde{p}_{j}})) = \\ = (\text{min}(\big \lfloor \frac{a_{\tilde{N}_{i}}}{n_{i}}\big \rfloor ,...,\big \lfloor \frac{a_{\tilde{N}_{j}}}{n_{j}}\big \rfloor );\text{min}(\big \lfloor \frac{m_{\tilde{N}_{i}}}{n_{i}}\big \rfloor ,...,\big \lfloor \frac{m_{\tilde{N}_{j}}}{n_{j}}\big \rfloor ); \text{min}(\big \lfloor \frac{b_{\tilde{N}_{i}}}{n_{i}}\big \rfloor ,...,\big \lfloor \frac{b_{\tilde{N}_{j}}}{n_{j}}\big \rfloor ))$
	\paragraph{Remainders:} $\\ \\ \tilde{N}_{i}^{'} = \tilde{N}_{i} - \tilde{p}n_{i} = (a_{\tilde{N}_{i}}; m_{\tilde{N}_{i}}; b_{\tilde{N}_{i}}) - (a_{\tilde{p}}n_{i}; m_{\tilde{p}}n_{i}; b_{\tilde{p}}n_{i}) = (a_{\tilde{N}_{i}} - b_{\tilde{p}}n_{i}; m_{\tilde{N}_{i}} - m_{\tilde{p}}n_{i}; b_{\tilde{N}_{i}} - a_{\tilde{p}}n_{i}) = \\ = (a_{\tilde{N}_{i}} - \text{min}(\big \lfloor \frac{b_{\tilde{N}_{i}}}{n_{i}}\big \rfloor ,...,\big \lfloor \frac{b_{\tilde{N}_{j}}}{n_{j}}\big \rfloor )n_{i}; m_{\tilde{N}_{i}} - \text{min}(\big \lfloor \frac{m_{\tilde{N}_{i}}}{n_{i}}\big \rfloor ,...,\big \lfloor \frac{m_{\tilde{N}_{j}}}{n_{j}}\big \rfloor )n_{i}; b_{\tilde{N}_{i}} - \text{min}(\big \lfloor \frac{a_{\tilde{N}_{i}}}{n_{i}}\big \rfloor ,...,\big \lfloor \frac{a_{\tilde{N}_{j}}}{n_{j}}\big \rfloor )n_{i})\\$
	$\dots$\\
	$\\ \tilde{N}_{j}^{'} = \tilde{N}_{j} - \tilde{p}n_{j} = (a_{\tilde{N}_{j}}; m_{\tilde{N}_{j}}; b_{\tilde{N}_{j}}) - (a_{\tilde{p}}n_{j}; m_{\tilde{p}}n_{j}; b_{\tilde{p}}n_{j}) = (a_{\tilde{N}_{j}} - b_{\tilde{p}}n_{j}; m_{\tilde{N}_{j}} - m_{\tilde{p}}n_{j}; b_{\tilde{N}_{j}} - a_{\tilde{p}}n_{j}) = \\ = (a_{\tilde{N}_{j}} - \text{min}(\big \lfloor \frac{b_{\tilde{N}_{i}}}{n_{i}}\big \rfloor ,...,\big \lfloor \frac{b_{\tilde{N}_{j}}}{n_{j}}\big \rfloor )n_{j}; m_{\tilde{N}_{j}} - \text{min}(\big \lfloor \frac{m_{\tilde{N}_{i}}}{n_{i}}\big \rfloor ,...,\big \lfloor \frac{m_{\tilde{N}_{j}}}{n_{j}}\big \rfloor )n_{j}; b_{\tilde{N}_{j}} - \text{min}(\big \lfloor \frac{a_{\tilde{N}_{i}}}{n_{i}}\big \rfloor ,...,\big \lfloor \frac{a_{\tilde{N}_{j}}}{n_{j}}\big \rfloor )n_{j})$
	\paragraph{Transformants:} $\\ \\ \tilde{q}_{k} = \tilde{p}r_{k} = (a_{\tilde{p}}r_{k}; m_{\tilde{p}}r_{k}; b_{\tilde{p}}r_{k}) = (\text{min}(\big \lfloor \frac{a_{\tilde{N}_{i}}}{n_{i}}\big \rfloor ,...,\big \lfloor \frac{a_{\tilde{N}_{j}}}{n_{j}}\big \rfloor )r_{k}; \text{min}(\big \lfloor \frac{m_{\tilde{N}_{i}}}{n_{i}}\big \rfloor ,...,\big \lfloor \frac{m_{\tilde{N}_{j}}}{n_{j}}\big \rfloor )r_{k}; \text{min}(\big \lfloor \frac{b_{\tilde{N}_{i}}}{n_{i}}\big \rfloor ,...,\big \lfloor \frac{b_{\tilde{N}_{j}}}{n_{j}}\big \rfloor )r_{k})\\$
	$\dots$
	$\\ \tilde{q}_{l} = \tilde{p}r_{l} = (a_{\tilde{p}}r_{l}; m_{\tilde{p}}r_{l}; b_{\tilde{p}}r_{l}) = (\text{min}(\big \lfloor \frac{a_{\tilde{N}_{i}}}{n_{i}}\big \rfloor ,...,\big \lfloor \frac{a_{\tilde{N}_{j}}}{n_{j}}\big \rfloor )r_{l}; \text{min}(\big \lfloor \frac{m_{\tilde{N}_{i}}}{n_{i}}\big \rfloor ,...,\big \lfloor \frac{m_{\tilde{N}_{j}}}{n_{j}}\big \rfloor )r_{l}; \text{min}(\big \lfloor \frac{b_{\tilde{N}_{i}}}{n_{i}}\big \rfloor ,...,\big \lfloor \frac{b_{\tilde{N}_{j}}}{n_{j}}\big \rfloor )r_{l})$
	\paragraph{Image cardinals:} $\\ \\ \tilde{N}_{k}^{'} = N_{k} + \tilde{q}_{k} = N_{k} + \tilde{p}r_{k} = (a_{\tilde{N}_{k}} + a_{\tilde{p}}r_{k}; m_{N_{k}} + m_{\tilde{p}}r_{k}; b_{N_{k}} + b_{\tilde{p}}r_{k}) = (a_{\tilde{N}_{k}} + \text{min}(\big \lfloor \frac{a_{\tilde{N}_{i}}}{n_{i}}\big \rfloor ,...,\big \lfloor \frac{a_{\tilde{N}_{j}}}{n_{j}}\big \rfloor )r_{k}; m_{\tilde{N}_{k}} + \text{min}(\big \lfloor \frac{m_{\tilde{N}_{i}}}{n_{i}}\big \rfloor ,...,\big \lfloor \frac{m_{\tilde{N}_{j}}}{n_{j}}\big \rfloor )r_{k}; b_{\tilde{N}_{k}} + \text{min}(\big \lfloor \frac{b_{\tilde{N}_{i}}}{n_{i}}\big \rfloor ,...,\big \lfloor \frac{b_{\tilde{N}_{j}}}{n_{j}}\big \rfloor )r_{k}))\\$
	$\dots$\\
	$\\ \tilde{N}_{l}^{'} = N_{l} + \tilde{q}_{l} = N_{l} + \tilde{p}r_{l} = (a_{\tilde{N}_{l}} + a_{\tilde{p}}r_{l}; m_{N_{l}} + m_{\tilde{p}}r_{l}; b_{N_{l}} + b_{\tilde{p}}r_{l}) = (a_{\tilde{N}_{l}} + \text{min}(\big \lfloor \frac{a_{\tilde{N}_{i}}}{n_{i}}\big \rfloor ,...,\big \lfloor \frac{a_{\tilde{N}_{j}}}{n_{j}}\big \rfloor )r_{l}; m_{\tilde{N}_{l}} + \text{min}(\big \lfloor \frac{m_{\tilde{N}_{i}}}{n_{i}}\big \rfloor ,...,\big \lfloor \frac{m_{\tilde{N}_{j}}}{n_{j}}\big \rfloor )r_{l}; b_{\tilde{N}_{l}} + \text{min}(\big \lfloor \frac{b_{\tilde{N}_{i}}}{n_{i}}\big \rfloor ,...,\big \lfloor \frac{b_{\tilde{N}_{j}}}{n_{j}}\big \rfloor )r_{l}))$
	\item \textbf{($N_{i},...,N_{j};\tilde{n}_{i},...,\tilde{n}_{j}; r_{k},...,r_{l}$)} - fuzzy radix
	\paragraph{Partial carries:} $\\ \\ \tilde{p}_{i} = (\big \lfloor \frac{N_{i}}{b_{\tilde{n}_{i}}}\big \rfloor ; \big \lfloor \frac{N_{i}}{m_{\tilde{n}_{i}}}\big \rfloor ; \big \lfloor \frac{N_{i}}{a_{\tilde{n}_{i}}}\big \rfloor )\\$
	$\dots$
	$\\ \tilde{p}_{j} = (\big \lfloor \frac{N_{j}}{b_{\tilde{n}_{j}}}\big \rfloor ; \big \lfloor \frac{N_{j}}{m_{\tilde{n}_{j}}}\big \rfloor ; \big \lfloor \frac{N_{j}}{a_{\tilde{n}_{j}}}\big \rfloor )$
	\paragraph{Fuzzy common carry:} $\\ \\ \tilde{p} = (\text{min}(a_{\tilde{p}_{i}},...,a_{\tilde{p}_{j}}); \text{min}(m_{\tilde{p}_{i}},...,m_{\tilde{p}_{j}}); \text{min}(b_{\tilde{p}_{i}},...b_{\tilde{p}_{j}})) =\\= (\text{min}(\big \lfloor \frac{N_{i}}{b_{\tilde{n}_{i}}}\big \rfloor ,...,\big \lfloor \frac{N_{j}}{b_{\tilde{n}_{j}}}\big \rfloor ); \text{min}(\big \lfloor \frac{N_{i}}{m_{\tilde{n}_{i}}}\big \rfloor ,...,\big \lfloor \frac{N_{j}}{m_{\tilde{n}_{j}}}\big \rfloor ); \text{min}(\big \lfloor \frac{N_{i}}{a_{\tilde{n}_{i}}}\big \rfloor ,...,\big \lfloor \frac{N_{j}}{a_{\tilde{n}_{j}}}\big \rfloor ))$
	\paragraph{Remainders:} $\\ \\ \tilde{N}_{i}^{'} = N_{i} - \tilde{p}\tilde{n}_{i} = N_{i} - (a_{\tilde{n}_{i}}a_{\tilde{p}}; m_{\tilde{n}_{i}}m_{\tilde{p}}; b_{\tilde{n}_{i}}b_{\tilde{p}}) = (N_{i} - b_{\tilde{n}_{i}}b_{\tilde{p}}; N_{i} - m_{\tilde{n}_{i}}m_{\tilde{p}}; N_{i} - a_{\tilde{n}_{i}}a_{\tilde{p}}) =\\= (N_{i} - b_{\tilde{n}_{i}}\text{min}(b_{\tilde{p}_{i}},...b_{\tilde{p}_{j}}); N_{i} - m_{\tilde{n}_{i}}\text{min}(m_{\tilde{p}_{i}},...,m_{\tilde{p}_{j}}); N_{i} - a_{\tilde{n}_{i}}\text{min}(a_{\tilde{p}_{i}},...,a_{\tilde{p}_{j}})\\$
	$\dots$\\
	$\\ \tilde{N}_{j}^{'} = N_{j} - \tilde{p}\tilde{n}_{j} = N_{j} - (a_{\tilde{n}_{j}}a_{\tilde{p}}; m_{\tilde{n}_{j}}m_{\tilde{p}}; b_{\tilde{n}_{j}}b_{\tilde{p}}) = (N_{j} - b_{\tilde{n}_{j}}b_{\tilde{p}}; N_{j} - m_{\tilde{n}_{j}}m_{\tilde{p}}; N_{j} - a_{\tilde{n}_{j}}a_{\tilde{p}}) =\\= (N_{j} - b_{\tilde{n}_{j}}\text{min}(b_{\tilde{p}_{i}},...b_{\tilde{p}_{j}}); N_{j} - m_{\tilde{n}_{j}}\text{min}(m_{\tilde{p}_{i}},...,m_{\tilde{p}_{j}}); N_{j} - a_{\tilde{n}_{j}}\text{min}(a_{\tilde{p}_{i}},...,a_{\tilde{p}_{j}})$
	\paragraph{Transformants:} $\\ \\ \tilde{q}_{k} = \tilde{p}r_{k} = (a_{\tilde{p}}r_{k}; m_{\tilde{p}}r_{k}; b_{\tilde{p}}r_{k}) = (\text{min}(a_{\tilde{p}_{i}},...,a_{\tilde{p}_{j}})r_{k}; \text{min}(m_{\tilde{p}_{i}},...,m_{\tilde{p}_{j}})r_{k}; \text{min}(b_{\tilde{p}_{i}},...b_{\tilde{p}_{j}})r_{k})\\$
	$\dots$
	$\\ \tilde{q}_{l} = \tilde{p}r_{l} = (a_{\tilde{p}}r_{l}; m_{\tilde{p}}r_{l}; b_{\tilde{p}}r_{l}) = (\text{min}(a_{\tilde{p}_{i}},...,a_{\tilde{p}_{j}})r_{l}; \text{min}(m_{\tilde{p}_{i}},...,m_{\tilde{p}_{j}})r_{l}; \text{min}(b_{\tilde{p}_{i}},...b_{\tilde{p}_{j}})r_{l})$
	\paragraph{Image cardinals:} $\\ \\ N_{k}^{'} = N_{k} + \tilde{q}_{k} = N_{k} + \tilde{p}r_{k} = (a_{\tilde{N}_{k}} + a_{\tilde{p}}r_{k}; m_{N_{k}} + m_{\tilde{p}}r_{k}; b_{N_{k}} + b_{\tilde{p}}r_{k}) = (a_{\tilde{N}_{k}} + \text{min}(\big \lfloor \frac{a_{\tilde{N}_{i}}}{n_{i}}\big \rfloor ,...,\big \lfloor \frac{a_{\tilde{N}_{j}}}{n_{j}}\big \rfloor )r_{k}; m_{\tilde{N}_{k}} + \text{min}(\big \lfloor \frac{m_{\tilde{N}_{i}}}{n_{i}}\big \rfloor ,...,\big \lfloor \frac{m_{\tilde{N}_{j}}}{n_{j}}\big \rfloor )r_{k}; b_{\tilde{N}_{k}} + \text{min}(\big \lfloor \frac{b_{\tilde{N}_{i}}}{n_{i}}\big \rfloor ,...,\big \lfloor \frac{b_{\tilde{N}_{j}}}{n_{j}}\big \rfloor )r_{k}))\\$
	$\dots$\\
	$\\ N_{l}^{'} = N_{l} + \tilde{q}_{l} = N_{l} + \tilde{p}r_{l} = (a_{\tilde{N}_{l}} + a_{\tilde{p}}r_{l}; m_{N_{l}} + m_{\tilde{p}}r_{l}; b_{N_{l}} + b_{\tilde{p}}r_{l}) = (a_{\tilde{N}_{l}} + \text{min}(\big \lfloor \frac{a_{\tilde{N}_{i}}}{n_{i}}\big \rfloor ,...,\big \lfloor \frac{a_{\tilde{N}_{j}}}{n_{j}}\big \rfloor )r_{l}; m_{\tilde{N}_{l}} + \text{min}(\big \lfloor \frac{m_{\tilde{N}_{i}}}{n_{i}}\big \rfloor ,...,\big \lfloor \frac{m_{\tilde{N}_{j}}}{n_{j}}\big \rfloor )r_{l}; b_{\tilde{N}_{l}} + \text{min}(\big \lfloor \frac{b_{\tilde{N}_{i}}}{n_{i}}\big \rfloor ,...,\big \lfloor \frac{b_{\tilde{N}_{j}}}{n_{j}}\big \rfloor )r_{l}))$
	\item \textbf{($N_{i},...,N_{j}; n_{i},...,n_{j}; \tilde{r}_{k},...,\tilde{r}_{l}$)} - fuzzy conversion rate
	\paragraph{Partial carries:} $\\ \\ p_{i} = \big \lfloor \frac{N_{i}}{n_{i}}\big \rfloor \\$
	$\dots$
	$\\ p_{j} = \big \lfloor \frac{N_{j}}{n_{j}}\big \rfloor $
	\paragraph{Fuzzy common carry:} $\\ \\ p = \text{min}(p_{i},...,p_{j}) = \text{min}(\big \lfloor \frac{N_{i}}{n_{i}}\big \rfloor ,...,\big \lfloor \frac{N_{j}}{n_{j}}\big \rfloor )$
	\paragraph{Remainders:} $\\ \\ N_{i}^{'} = N_{i} - pn_{i} = N_{i} - \text{min}(\big \lfloor \frac{N_{i}}{n_{i}}\big \rfloor ,...,\big \lfloor \frac{N_{j}}{n_{j}}\big \rfloor )n_{i}\\$
	$\dots$\\
	$\\ N_{j}^{'} = N_{j} - pn_{j} = N_{j} - \text{min}(\big \lfloor \frac{N_{i}}{n_{i}}\big \rfloor ,...,\big \lfloor \frac{N_{j}}{n_{j}}\big \rfloor )n_{j}$
	\paragraph{Transformants:} $\\ \\ \tilde{q}_{k} = p\tilde{r}_{k} = (pa_{\tilde{r}_{k}}; pm_{\tilde{r}_{k}}; pb_{\tilde{r}_{k}}) = (\text{min}(\big \lfloor \frac{N_{i}}{n_{i}}\big \rfloor ,...,\big \lfloor \frac{N_{j}}{n_{j}}\big \rfloor )a_{\tilde{r}_{k}}; \text{min}(\big \lfloor \frac{N_{i}}{n_{i}}\big \rfloor ,...,\big \lfloor \frac{N_{j}}{n_{j}}\big \rfloor )m_{\tilde{r}_{k}}; \text{min}(\big \lfloor \frac{N_{i}}{n_{i}}\big \rfloor ,...,\big \lfloor \frac{N_{j}}{n_{j}}\big \rfloor )b_{\tilde{r}_{k}})\\$
	$\dots$
	$\\ \tilde{q}_{l} = p\tilde{r}_{l} = (pa_{\tilde{r}_{l}}; pm_{\tilde{r}_{l}}; pb_{\tilde{r}_{l}}) = (\text{min}(\big \lfloor \frac{N_{i}}{n_{i}}\big \rfloor ,...,\big \lfloor \frac{N_{j}}{n_{j}}\big \rfloor )a_{\tilde{r}_{l}}; \text{min}(\big \lfloor \frac{N_{i}}{n_{i}}\big \rfloor ,...,\big \lfloor \frac{N_{j}}{n_{j}}\big \rfloor )m_{\tilde{r}_{l}}; \text{min}(\big \lfloor \frac{N_{i}}{n_{i}}\big \rfloor ,...,\big \lfloor \frac{N_{j}}{n_{j}}\big \rfloor )b_{\tilde{r}_{l}})$
	\paragraph{Image cardinals:} $\\ \\ \tilde{N}_{k}^{'} = N_{k} + \tilde{q}_{k} = N_{k} + p\tilde{r}_{k} = (N_{k} + \text{min}(\big \lfloor \frac{N_{i}}{n_{i}}\big \rfloor ,...,\big \lfloor \frac{N_{j}}{n_{j}}\big \rfloor )a_{\tilde{r}_{k}}; N_{k} + \text{min}(\big \lfloor \frac{N_{i}}{n_{i}}\big \rfloor ,...,\big \lfloor \frac{N_{j}}{n_{j}}\big \rfloor )m_{\tilde{r}_{k}}; N_{k} + \text{min}(\big \lfloor \frac{N_{i}}{n_{i}}\big \rfloor ,...,\big \lfloor \frac{N_{j}}{n_{j}}\big \rfloor )b_{\tilde{r}_{k}})\\$
	$\dots$\\
	$\\ \tilde{N}_{l}^{'} = N_{l} + \tilde{q}_{k} = N_{l} + p\tilde{r}_{l} = (N_{l} + \text{min}(\big \lfloor \frac{N_{i}}{n_{i}}\big \rfloor ,...,\big \lfloor \frac{N_{j}}{n_{j}}\big \rfloor )a_{\tilde{r}_{l}}; N_{l} + \text{min}(\big \lfloor \frac{N_{i}}{n_{i}}\big \rfloor ,...,\big \lfloor \frac{N_{j}}{n_{j}}\big \rfloor )m_{\tilde{r}_{l}}; N_{l} + \text{min}(\big \lfloor \frac{N_{i}}{n_{i}}\big \rfloor ,...,\big \lfloor \frac{N_{j}}{n_{j}}\big \rfloor )b_{\tilde{r}_{l}})$
	\item \textbf{($N_{i},...,N_{j}; \tilde{n}_{i},...,\tilde{n}_{j}; \tilde{r}_{k},...,\tilde{r}_{l}$)} -fuzzy radix and conversion rate
	\paragraph{Partial carries:} $\\ \\ \tilde{p}_{i} = \big \lfloor \frac{N_{i}}{\tilde{n}_{i}}\big \rfloor  = (\big \lfloor \frac{N_{i}}{b_{\tilde{n}_{i}}}\big \rfloor ; \big \lfloor \frac{N_{i}}{m_{\tilde{n}_{i}}}\big \rfloor ; \big \lfloor \frac{N_{i}}{a_{\tilde{n}_{i}}}\big \rfloor )\\$
	$\dots$
	$\\ \tilde{p}_{j} = \big \lfloor \frac{N_{j}}{\tilde{n}_{j}}\big \rfloor  = (\big \lfloor \frac{N_{j}}{b_{\tilde{n}_{j}}}\big \rfloor ; \big \lfloor \frac{N_{j}}{m_{\tilde{n}_{j}}}\big \rfloor ; \big \lfloor \frac{N_{j}}{a_{\tilde{n}_{j}}}\big \rfloor )$
	\paragraph{Fuzzy common carry:} $\\ \\ \tilde{p} = (\text{min}(a_{\tilde{p}_{i}},...,a_{\tilde{p}_{j}}); \text{min}(m_{\tilde{p}_{i}},...,m_{\tilde{p}_{j}}); \text{min}(b_{\tilde{p}_{i}},...b_{\tilde{p}_{j}})) = \\ = (\text{min}(\big \lfloor \frac{N_{i}}{b_{\tilde{n}_{i}}}\big \rfloor ,...,\big \lfloor \frac{N_{j}}{b_{\tilde{n}_{j}}}\big \rfloor ); \text{min}(\big \lfloor \frac{N_{i}}{m_{\tilde{n}_{i}}}\big \rfloor ,...,\big \lfloor \frac{N_{i}}{m_{\tilde{n}_{j}}}\big \rfloor ); \text{min}(\big \lfloor \frac{N_{i}}{a_{\tilde{n}_{i}}}\big \rfloor ,...,\big \lfloor \frac{N_{i}}{a_{\tilde{n}_{j}}}\big \rfloor ))$
	\paragraph{Remainders:} $\\ \\ \tilde{N}_{i}^{'} = N_{i} - \tilde{p}\tilde{n}_{i} = N_{i} - (a_{\tilde{p}}a_{\tilde{n}_{i}}; m_{\tilde{p}}m_{\tilde{n}_{i}}; b_{\tilde{p}}b_{\tilde{n}_{i}}) = (N_{i} - b_{\tilde{p}}b_{\tilde{n}_{i}}; N_{i} - m_{\tilde{p}}m_{\tilde{n}_{i}}; N_{i} - a_{\tilde{p}}a_{\tilde{n}_{i}}) = \\ = (N_{i} - \text{min}(b_{\tilde{p}_{i}},...b_{\tilde{p}_{j}})b_{\tilde{n}_{i}}; N_{i} - \text{min}(m_{\tilde{p}_{i}},...,m_{\tilde{p}_{j}})m_{\tilde{n}_{i}}; N_{i} - \text{min}(a_{\tilde{p}_{i}},...,a_{\tilde{p}_{j}})a_{\tilde{n}_{i}}) =\\=(N_{i} - \text{min}(\big \lfloor \frac{N_{i}}{a_{\tilde{n}_{i}}}\big \rfloor ,...,\big \lfloor \frac{N_{i}}{a_{\tilde{n}_{j}}}\big \rfloor )b_{\tilde{n}_{i}}; N_{i} - \text{min}(\big \lfloor \frac{N_{i}}{m_{\tilde{n}_{i}}}\big \rfloor ,...,\big \lfloor \frac{N_{i}}{m_{\tilde{n}_{j}}}\big \rfloor )m_{\tilde{n}_{i}}; N_{i} - \text{min}(\big \lfloor \frac{N_{i}}{b_{\tilde{n}_{i}}}\big \rfloor ,...,\big \lfloor \frac{N_{j}}{b_{\tilde{n}_{j}}}\big \rfloor )a_{\tilde{n}_{i}})\\$
	$\dots$\\
	$\\ \tilde{N}_{j}^{'} = N_{j} - \tilde{p}\tilde{n}_{j} = N_{j} - (a_{\tilde{p}}a_{\tilde{n}_{j}}; m_{\tilde{p}}m_{\tilde{n}_{j}}; b_{\tilde{p}}b_{\tilde{n}_{j}}) = (N_{j} - b_{\tilde{p}}b_{\tilde{n}_{j}}; N_{j} - m_{\tilde{p}}m_{\tilde{n}_{j}}; N_{j} - a_{\tilde{p}}a_{\tilde{n}_{j}}) = \\ = (N_{j} - \text{min}(b_{\tilde{p}_{i}},...b_{\tilde{p}_{j}})b_{\tilde{n}_{j}}; N_{j} - \text{min}(m_{\tilde{p}_{i}},...,m_{\tilde{p}_{j}})m_{\tilde{n}_{j}}; N_{j} - \text{min}(a_{\tilde{p}_{i}},...,a_{\tilde{p}_{j}})a_{\tilde{n}_{j}}) =\\=(N_{j} - \text{min}(\big \lfloor \frac{N_{i}}{a_{\tilde{n}_{i}}}\big \rfloor ,...,\big \lfloor \frac{N_{i}}{a_{\tilde{n}_{j}}}\big \rfloor )b_{\tilde{n}_{j}}; N_{j} - \text{min}(\big \lfloor \frac{N_{i}}{m_{\tilde{n}_{i}}}\big \rfloor ,...,\big \lfloor \frac{N_{i}}{m_{\tilde{n}_{j}}}\big \rfloor )m_{\tilde{n}_{j}}; N_{j} - \text{min}(\big \lfloor \frac{N_{i}}{b_{\tilde{n}_{i}}}\big \rfloor ,...,\big \lfloor \frac{N_{j}}{b_{\tilde{n}_{j}}}\big \rfloor )a_{\tilde{n}_{j}})$
	\paragraph{Transformants:} $\\ \\ \tilde{q}_{k} = \tilde{p}\tilde{r}_{k} = (a_{\tilde{p}}a_{\tilde{r}_{k}}; m_{\tilde{p}}m_{\tilde{r}_{k}}; b_{\tilde{p}}b_{\tilde{r}_{k}}) = (\text{min}(a_{\tilde{p}_{i}},...,a_{\tilde{p}_{j}})a_{\tilde{r}_{k}}; \text{min}(m_{\tilde{p}_{i}},...,m_{\tilde{p}_{j}})m_{\tilde{r}_{k}}; \text{min}(b_{\tilde{p}_{i}},...b_{\tilde{p}_{j}})b_{\tilde{r}_{k}}) =\\= (\text{min}(\big \lfloor \frac{N_{i}}{b_{\tilde{n}_{i}}}\big \rfloor ,...,\big \lfloor \frac{N_{j}}{b_{\tilde{n}_{j}}}\big \rfloor )a_{\tilde{r}_{k}}; \text{min}(\big \lfloor \frac{N_{i}}{m_{\tilde{n}_{i}}}\big \rfloor ,...,\big \lfloor \frac{N_{i}}{m_{\tilde{n}_{j}}}\big \rfloor )m_{\tilde{r}_{k}}; \text{min}(\big \lfloor \frac{N_{i}}{a_{\tilde{n}_{i}}}\big \rfloor ,...,\big \lfloor \frac{N_{i}}{a_{\tilde{n}_{j}}}\big \rfloor )b_{\tilde{r}_{k}})\\$
	$\dots$\\
	$\\ \tilde{q}_{l} = \tilde{p}\tilde{r}_{l} = (a_{\tilde{p}}a_{\tilde{r}_{l}}; m_{\tilde{p}}m_{\tilde{r}_{l}}; b_{\tilde{p}}b_{\tilde{r}_{l}}) = (\text{min}(a_{\tilde{p}_{i}},...,a_{\tilde{p}_{j}})a_{\tilde{r}_{l}}; \text{min}(m_{\tilde{p}_{i}},...,m_{\tilde{p}_{j}})m_{\tilde{r}_{l}}; \text{min}(b_{\tilde{p}_{i}},...b_{\tilde{p}_{j}})b_{\tilde{r}_{l}}) =\\= (\text{min}(\big \lfloor \frac{N_{i}}{b_{\tilde{n}_{i}}}\big \rfloor ,...,\big \lfloor \frac{N_{j}}{b_{\tilde{n}_{j}}}\big \rfloor )a_{\tilde{r}_{l}}; \text{min}(\big \lfloor \frac{N_{i}}{m_{\tilde{n}_{i}}}\big \rfloor ,...,\big \lfloor \frac{N_{i}}{m_{\tilde{n}_{j}}}\big \rfloor )m_{\tilde{r}_{l}}; \text{min}(\big \lfloor \frac{N_{i}}{a_{\tilde{n}_{i}}}\big \rfloor ,...,\big \lfloor \frac{N_{i}}{a_{\tilde{n}_{j}}}\big \rfloor )b_{\tilde{r}_{l}})$
	\paragraph{Image cardinals:} $\\ \\ \tilde{N}_{k}^{'} = N_{k} + \tilde{q}_{k} = N_{k} + \tilde{p}\tilde{r}_{k} = (N_{k} + a_{\tilde{p}}a_{\tilde{r}_{k}}; N_{k} + m_{\tilde{p}}m_{\tilde{r}_{k}}; N_{k} + b_{\tilde{p}}b_{\tilde{r}_{k}}) = \\ = (N_{k} + \text{min}(a_{\tilde{p}_{i}},...,a_{\tilde{p}_{j}})a_{\tilde{r}_{k}}; N_{k} + \text{min}(m_{\tilde{p}_{i}},...,m_{\tilde{p}_{j}})m_{\tilde{r}_{k}}; N_{k} +\text{min}(b_{\tilde{p}_{i}},...b_{\tilde{p}_{j}})b_{\tilde{r}_{k}}) = \\ = (N_{k} + \text{min}(\big \lfloor \frac{N_{i}}{b_{\tilde{n}_{i}}}\big \rfloor ,...,\big \lfloor \frac{N_{j}}{b_{\tilde{n}_{j}}}\big \rfloor )a_{\tilde{r}_{k}}; N_{k} + \text{min}(\big \lfloor \frac{N_{i}}{m_{\tilde{n}_{i}}}\big \rfloor ,...,\big \lfloor \frac{N_{i}}{m_{\tilde{n}_{j}}}\big \rfloor )m_{\tilde{r}_{k}}; N_{k} +\text{min}(\big \lfloor \frac{N_{i}}{a_{\tilde{n}_{i}}}\big \rfloor ,...,\big \lfloor \frac{N_{j}}{a_{\tilde{n}_{j}}}\big \rfloor )b_{\tilde{r}_{k}})\\$
	$\dots$\\
	$\\ \tilde{N}_{l}^{'} = N_{l} + \tilde{q}_{l} = N_{l} + \tilde{p}\tilde{r}_{l} = (N_{l} + a_{\tilde{p}}a_{\tilde{r}_{l}}; N_{l} + m_{\tilde{p}}m_{\tilde{r}_{l}}; N_{l} + b_{\tilde{p}}b_{\tilde{r}_{l}}) = \\ = (N_{l} + \text{min}(a_{\tilde{p}_{i}},...,a_{\tilde{p}_{j}})a_{\tilde{r}_{l}}; N_{l} + \text{min}(m_{\tilde{p}_{i}},...,m_{\tilde{p}_{j}})m_{\tilde{r}_{l}}; N_{l} +\text{min}(b_{\tilde{p}_{i}},...b_{\tilde{p}_{j}})b_{\tilde{r}_{l}}) = \\ = (N_{l} + \text{min}(\big \lfloor \frac{N_{i}}{b_{\tilde{n}_{i}}}\big \rfloor ,...,\big \lfloor \frac{N_{j}}{b_{\tilde{n}_{j}}}\big \rfloor )a_{\tilde{r}_{l}}; N_{l} + \text{min}(\big \lfloor \frac{N_{i}}{m_{\tilde{n}_{i}}}\big \rfloor ,...,\big \lfloor \frac{N_{i}}{m_{\tilde{n}_{j}}}\big \rfloor )m_{\tilde{r}_{l}}; N_{l} +\text{min}(\big \lfloor \frac{N_{i}}{a_{\tilde{n}_{i}}}\big \rfloor ,...,\big \lfloor \frac{N_{j}}{a_{\tilde{n}_{j}}}\big \rfloor )b_{\tilde{r}_{l}})$
	\item \textbf{($\tilde{N}_{i},...,\tilde{N}_{j}; \tilde{n}_{i},...,\tilde{n}_{j}; \tilde{r}_{k},...,\tilde{r}_{l}$)} - whole fuzziness
	\paragraph{Partial carries:} $\\ \\ \tilde{p}_{i} = (\big \lfloor \frac{a_{\tilde{N}_{i}}}{b_{\tilde{n}_{i}}}\big \rfloor ; \big \lfloor \frac{m_{\tilde{N}_{i}}}{m_{\tilde{n}_{i}}}\big \rfloor ; \big \lfloor \frac{b_{\tilde{N}_{i}}}{a_{\tilde{n}_{i}}}\big \rfloor )\\$
	$\dots$
	$\\ \tilde{p}_{j} = (\big \lfloor \frac{a_{\tilde{N}_{j}}}{b_{\tilde{n}_{j}}}\big \rfloor ; \big \lfloor \frac{m_{\tilde{N}_{j}}}{m_{\tilde{n}_{j}}}\big \rfloor ; \big \lfloor \frac{b_{\tilde{N}_{j}}}{a_{\tilde{n}_{j}}}\big \rfloor )$
	\paragraph{Fuzzy common carry:} $\\ \\ \tilde{p} = (\text{min}(a_{\tilde{p}_{i}},...,a_{\tilde{p}_{j}}); \text{min}(m_{\tilde{p}_{i}},...,m_{\tilde{p}_{j}}); \text{min}(b_{\tilde{p}_{i}},...b_{\tilde{p}_{j}})) =\\= (\text{min}(\big \lfloor \frac{a_{\tilde{N}_{i}}}{b_{\tilde{n}_{i}}}\big \rfloor ,...,\big \lfloor \frac{a_{\tilde{N}_{j}}}{b_{\tilde{n}_{j}}}\big \rfloor ); \text{min}(\big \lfloor \frac{m_{\tilde{N}_{i}}}{m_{\tilde{n}_{i}}}\big \rfloor ,...,\big \lfloor \frac{m_{\tilde{N}_{j}}}{m_{\tilde{n}_{j}}}\big \rfloor ); \text{min}(\big \lfloor \frac{b_{\tilde{N}_{i}}}{a_{\tilde{n}_{i}}}\big \rfloor ,...,\big \lfloor \frac{b_{\tilde{N}_{j}}}{a_{\tilde{n}_{j}}}\big \rfloor ))$
	\paragraph{Remainders:} $\\ \\ \tilde{N}_{i}^{'} = \tilde{N}_{i} - \tilde{p}\tilde{n}_{i} = (a_{\tilde{N}_{i}}; m_{\tilde{N}_{i}}; b_{\tilde{N}_{i}}) - (a_{\tilde{p}}a_{\tilde{n}_{i}}; m_{\tilde{p}}m_{\tilde{n}_{i}}; b_{\tilde{p}}b_{\tilde{n}_{i}}) = (a_{\tilde{N}_{i}} - b_{\tilde{p}}b_{\tilde{n}_{i}}; m_{\tilde{N}_{i}} - m_{\tilde{p}}m_{\tilde{n}_{i}}; b_{\tilde{N}_{i}} - a_{\tilde{p}}a_{\tilde{n}_{i}}) =\\= (a_{\tilde{N}_{i}} - \text{min}(\big \lfloor \frac{b_{\tilde{N}_{i}}}{a_{\tilde{n}_{i}}}\big \rfloor ,...,\big \lfloor \frac{b_{\tilde{N}_{j}}}{a_{\tilde{n}_{j}}}\big \rfloor )b_{\tilde{n}_{i}}; m_{\tilde{N}_{i}} - \text{min}(\big \lfloor \frac{m_{\tilde{N}_{i}}}{m_{\tilde{n}_{i}}}\big \rfloor ,...,\big \lfloor \frac{m_{\tilde{N}_{j}}}{m_{\tilde{n}_{j}}}\big \rfloor )m_{\tilde{n}_{i}}; b_{\tilde{N}_{i}} - \text{min}(\big \lfloor \frac{a_{\tilde{N}_{i}}}{b_{\tilde{n}_{i}}}\big \rfloor ,...,\big \lfloor \frac{a_{\tilde{N}_{j}}}{b_{\tilde{n}_{j}}}\big \rfloor )a_{\tilde{n}_{i}})\\$
	$\dots$\\
	$\\ \tilde{N}_{j}^{'} = \tilde{N}_{j} - \tilde{p}\tilde{n}_{j} = (a_{\tilde{N}_{j}}; m_{\tilde{N}_{j}}; b_{\tilde{N}_{j}}) - (a_{\tilde{p}}a_{\tilde{n}_{j}}; m_{\tilde{p}}m_{\tilde{n}_{j}}; b_{\tilde{p}}b_{\tilde{n}_{j}}) = (a_{\tilde{N}_{j}} - b_{\tilde{p}}b_{\tilde{n}_{j}}; m_{\tilde{N}_{j}} - m_{\tilde{p}}m_{\tilde{n}_{j}}; b_{\tilde{N}_{j}} - a_{\tilde{p}}a_{\tilde{n}_{j}}) =\\= (a_{\tilde{N}_{j}} - \text{min}(\big \lfloor \frac{b_{\tilde{N}_{i}}}{a_{\tilde{n}_{i}}}\big \rfloor ,...,\big \lfloor \frac{b_{\tilde{N}_{j}}}{a_{\tilde{n}_{j}}}\big \rfloor )b_{\tilde{n}_{j}}; m_{\tilde{N}_{j}} - \text{min}(\big \lfloor \frac{m_{\tilde{N}_{i}}}{m_{\tilde{n}_{i}}}\big \rfloor ,...,\big \lfloor \frac{m_{\tilde{N}_{j}}}{m_{\tilde{n}_{j}}}\big \rfloor )m_{\tilde{n}_{j}}; b_{\tilde{N}_{j}} - \text{min}(\big \lfloor \frac{a_{\tilde{N}_{i}}}{b_{\tilde{n}_{i}}}\big \rfloor ,...,\big \lfloor \frac{a_{\tilde{N}_{j}}}{b_{\tilde{n}_{j}}}\big \rfloor )a_{\tilde{n}_{j}})$
	\paragraph{Transformants:} $\\ \\ \tilde{q}_{k} = \tilde{p}\tilde{r}_{k} = (a_{\tilde{p}}a_{\tilde{r}_{k}}; m_{\tilde{p}}m_{\tilde{r}_{k}}; b_{\tilde{p}}b_{\tilde{r}_{k}}) = (\text{min}(\big \lfloor \frac{a_{\tilde{N}_{i}}}{b_{\tilde{n}_{i}}}\big \rfloor ,...,\big \lfloor \frac{a_{\tilde{N}_{j}}}{b_{\tilde{n}_{j}}}\big \rfloor )a_{\tilde{r}_{k}}; \text{min}(\big \lfloor \frac{m_{\tilde{N}_{i}}}{m_{\tilde{n}_{i}}}\big \rfloor ,...,\big \lfloor \frac{m_{\tilde{N}_{j}}}{m_{\tilde{n}_{j}}}\big \rfloor )m_{\tilde{r}_{k}};\text{min}(\big \lfloor \frac{b_{\tilde{N}_{i}}}{a_{\tilde{n}_{i}}}\big \rfloor ,...,\big \lfloor \frac{b_{\tilde{N}_{j}}}{a_{\tilde{n}_{j}}}\big \rfloor )b_{\tilde{r}_{k}})\\$
	$\dots$
	$\\ \tilde{q}_{l} = \tilde{p}\tilde{r}_{l} = (a_{\tilde{p}}a_{\tilde{r}_{l}}; m_{\tilde{p}}m_{\tilde{r}_{l}}; b_{\tilde{p}}b_{\tilde{r}_{l}}) = (\text{min}(\big \lfloor \frac{a_{\tilde{N}_{i}}}{b_{\tilde{n}_{i}}}\big \rfloor ,...,\big \lfloor \frac{a_{\tilde{N}_{j}}}{b_{\tilde{n}_{j}}}\big \rfloor )a_{\tilde{r}_{l}}; \text{min}(\big \lfloor \frac{m_{\tilde{N}_{i}}}{m_{\tilde{n}_{i}}}\big \rfloor ,...,\big \lfloor \frac{m_{\tilde{N}_{j}}}{m_{\tilde{n}_{j}}}\big \rfloor )m_{\tilde{r}_{l}};\text{min}(\big \lfloor \frac{b_{\tilde{N}_{i}}}{a_{\tilde{n}_{i}}}\big \rfloor ,...,\big \lfloor \frac{b_{\tilde{N}_{j}}}{a_{\tilde{n}_{j}}}\big \rfloor )b_{\tilde{r}_{l}})$
	\paragraph{Image cardinals:} $\\ \\ \tilde{N}_{k}^{'} = N_{l} + \tilde{q}_{k} = \tilde{N}_{k} + \tilde{p}\tilde{r}_{k} = (\tilde{N}_{k} + a_{\tilde{p}}a_{\tilde{r}_{k}}; \tilde{N}_{k} + m_{\tilde{p}}m_{\tilde{r}_{k}}; \tilde{N}_{k} + b_{\tilde{p}}b_{\tilde{r}_{k}}) =\\= (\tilde{N}_{k} + \text{min}(\big \lfloor \frac{a_{\tilde{N}_{i}}}{b_{\tilde{n}_{i}}}\big \rfloor ,...,\big \lfloor \frac{a_{\tilde{N}_{j}}}{b_{\tilde{n}_{j}}}\big \rfloor )a_{\tilde{r}_{k}}; \tilde{N}_{k} + \text{min}(\big \lfloor \frac{m_{\tilde{N}_{i}}}{m_{\tilde{n}_{i}}}\big \rfloor ,...,\big \lfloor \frac{m_{\tilde{N}_{j}}}{m_{\tilde{n}_{j}}}\big \rfloor )m_{\tilde{r}_{k}}; \tilde{N}_{k} + \text{min}(\big \lfloor \frac{b_{\tilde{N}_{i}}}{a_{\tilde{n}_{i}}}\big \rfloor ,...,\big \lfloor \frac{b_{\tilde{N}_{j}}}{a_{\tilde{n}_{j}}}\big \rfloor )b_{\tilde{r}_{k}})\\$
	$\dots$\\
	$\\ \tilde{N}_{l}^{'} = N_{l} + \tilde{q}_{l} = \tilde{N}_{l} + \tilde{p}\tilde{r}_{l} = (\tilde{N}_{l} + a_{\tilde{p}}a_{\tilde{r}_{l}}; \tilde{N}_{l} + m_{\tilde{p}}m_{\tilde{r}_{l}}; \tilde{N}_{l} + b_{\tilde{p}}b_{\tilde{r}_{l}}) =\\= (\tilde{N}_{l} + \text{min}(\big \lfloor \frac{a_{\tilde{N}_{i}}}{b_{\tilde{n}_{i}}}\big \rfloor ,...,\big \lfloor \frac{a_{\tilde{N}_{j}}}{b_{\tilde{n}_{j}}}\big \rfloor )a_{\tilde{r}_{l}}; \tilde{N}_{l} + \text{min}(\big \lfloor \frac{m_{\tilde{N}_{i}}}{m_{\tilde{n}_{i}}}\big \rfloor ,...,\big \lfloor \frac{m_{\tilde{N}_{j}}}{m_{\tilde{n}_{j}}}\big \rfloor )m_{\tilde{r}_{l}}; \tilde{N}_{l} + \text{min}(\big \lfloor \frac{b_{\tilde{N}_{i}}}{a_{\tilde{n}_{i}}}\big \rfloor ,...,\big \lfloor \frac{b_{\tilde{N}_{j}}}{a_{\tilde{n}_{j}}}\big \rfloor )b_{\tilde{r}_{l}})$
\end{enumerate}
\end{document}